\documentclass[10pt,twocolumn,letterpaper]{article}

\usepackage{bm}
\usepackage{algorithm}

\usepackage{tensor}
\usepackage[nocompress]{cite}
\usepackage[pdftex]{graphicx}
\usepackage[cmex10]{amsmath}
\usepackage{amssymb}
\usepackage{mathtools}
\usepackage{tabularx}

\usepackage{algorithmic}
\usepackage{array}

\usepackage{url}

\newcommand*{\Scale}[2]{\scalebox{#1}{$#2$}}%

\DeclareMathOperator{\tr}{tr}
  
\def\ve#1{\mathbf{#1}}

\def\inv#1{#1^{\mathclap{\substack{ \ \Scale{0.5}{-}1 \\ \  }}} \ }

\def\norm#1{\left|\left|{#1}\right|\right|}
\def\mCross#1{[{#1}]_{\times}}
\def\smalleq#1{\small{#1}\normalsize} 

\makeatletter
 
\def\ie{\emph{i.e.}}

\def\etal{\emph{et al.}}
\makeatother

\makeatletter
\newcommand*{\centerfloat}{%
  \parindent \z@
  \leftskip \z@ \@plus 1fil \@minus \textwidth
  \rightskip\leftskip
  \parfillskip \z@skip}
\makeatother

\begin{document}
%
\title{RGBiD-SLAM for Accurate Real-time Localisation and 3D Mapping}
%
%
%

\author{Daniel~Gutierrez-Gomez        
        and~Jose~J.~Guerrero
\thanks{D. Gutierrez-Gomez and J. J. Guerrero are with Instituto de Investigacion en Ingenieria de Aragon and Departamento de Informatica e Ingenieria de Sistemas, Universidad de Zaragoza, Maria de Luna 1, 500018, Zaragoza (Spain).
e-mail: .}
\thanks{This work has been supported by projects}
}

\maketitle

\begin{abstract}
In this paper we present a complete SLAM system for RGB-D cameras, namely RGB-iD SLAM. The presented approach is a dense 
direct SLAM method with the main characteristic of working with the depth maps in inverse depth parametrisation for the 
routines of dense alignment or keyframe fusion. The system consists in 2 CPU threads working in parallel, which share 
the use of the GPU for dense alignment and keyframe fusion routines. The first thread is a front-end operating at frame 
rate, which processes every incoming frame from the RGB-D sensor to compute the incremental odometry and integrate it in 
a keyframe which is changed periodically following a covisibility-based strategy. The second thread is a back-end which 
receives keyframes from the front-end. This thread is in charge of segmenting the keyframes based on their structure, 
describing them using Bags of Words, trying to find potential loop closures with previous keyframes, and in such case 
perform pose-graph optimisation for trajectory correction. In 
addition, our system allows is able to compute the odometry both with unregistered and registered depth maps, allowing to use customised calibrations of the RGB-D sensor. As a consequence in the paper we also propose a detailed calibration pipeline to compute customised calibrations for particular RGB-D cameras. The experiments with our approach in the TUM RGB-D benchmark datasets show results superior in accuracy to the state-of-the-art in many of the sequences. The code has been made available on-line for research purposes \footnote{\texttt{https://github.com/dangut/RGBiD-SLAM}}.
\end{abstract}

\section{Introduction}

In the last years visual SLAM has become one fertile research topic in the fields of computer vision and robotics. Visual SLAM can be addressed either by feature-based or direct feature-less methods. In a previous step feature-based methods extract pixel locations with high gradients and compute the motion by minimisation of the pixelic reprojection error. Feature-less or direct methods skip this step and compute the motion estimate directly from the raw camera input. Regardless which method is used a complete visual SLAM system usually consists of the following modules:

\begin{itemize}
 \item {Camera tracking}
 \item {3D reconstruction}
 \item {Place recognition and loop closure when a zone is revisited}
\end{itemize}

One of the earliest and most influential works on real time monocular SLAM was presented by Davison \cite{Davison2003}. It is a visual feature-based SLAM system, where the camera motion and position of landmarks of the environment are estimated using an Extended Kalman Filter (EKF) from the image displacement of the image projections, or features, of such landmarks. Following the success of this work, several contributions introduced new improvements and functionalities like the inverse depth parametrisation of the landmarks \cite{Civera2008}, a relocalisation module \cite{Williams2007}, and efficient scheme for data association and outlier rejection \cite{Civera2010} or a generalised projection model to support catadioptric sensors \cite{GutierrezGomez2011}. 

One weakness of this approach is that following a probabilistic filter paradigm, camera tracking and mapping are performed both at frame rate, which limits severally the number of landmarks which could be reconstructed. Klein and Murray \cite{Klein2007} addressed this issue with their PTAM (Parallel Tracking and Mapping), where camera tracking and mapping are separated into two different threads. The tracking thread operates at frame rate and is on charge of estimating the camera position given a fixed map; while the mapping thread operates at a lower rate, by building or updating a map by performing bundle adjustment between a set of keyframes selected from the total number of frames. Mur-Artal et al. presented in \cite{MurArtal2015} a visual SLAM approach addressing the weaknesses of PTAM, like handling of landmarks occlusions or mapping scalability, as well as including a novel loop closure and relocalisation method based on Bags of Binary Words using ORB descriptors.

Compared to feature-based methods like the aforementioned direct methods have the advantage of skipping the feature extraction step which performed only an a small fraction of the image pixels and is frequently based on heuristics. However feature-less or direct monocular SLAM methods are more challenging due to the lack of depth measurements from vision sensors and the high computational burden of performing per-pixel operations. 

Maybe the earliest and most influencing work in real-time dense monocular method was DTAM (Dense Tracking and Mapping) 
presented by Newcombe et al. \cite{DTAM}. The authors propose to reconstruct the depth of the scene in keyframes and at 
the same time  estimate the pose of the camera. The problem of estimating a dense depth map from just RGB images is 
ill-posed and thus authors used an optimisation framework based on variational methods and implemented on GPU. In 
\cite{Engel2013}, Engel et al. propose a direct monocular SLAM method which obtains a semi-dense mapping based on 
variable baseline matching.

Nevertheless, in the last years visual SLAM has been very influenced by two differential events which have rendered direct feature-free approaches more tractable. First the development of a new generation of GPUs and CPUs, which allowed to design algorithms with high parallelisation; and secondly the advent of depth sensors, which are able to provide a dense depth image of the observed environment. The earlier prompted the development of real-time dense visual SLAM methods, while the later enabled SLAM, both feature-based and direct, using RGB and depth channels from a RGB-D camera.

Though RGB-D cameras have the great advantage of providing dense depth information, with respect to RGB cameras they 
present a more complex calibration. To perform localisation and mapping with RGB, one only needs to calibrate a 
conventional camera which nowadays is a relatively simple process by using for example Bouguet calibration toolbox 
\cite{BouguetCalibration}. On the other hand, RGB-D sensors require a stereo calibration between the RGB and depth 
cameras for depth registration. In addition to that, since the depth sensor is really measuring a pixel disparity and 
then converts it to depth, this relation has to be calibrated also. RGB-D sensors have this functions hard-coded in the 
hardware, being able depth computation and registration internally. However the use of a fixed factory calibration does 
not account for imperfections of particular sensors and a customised calibration might be necessary to reach good 
accuracy in SLAM with RGB-D sensors. 

In this paper we present a complete SLAM system for custom calibrated RGB-D sensors. The developed camera tracking module builds upon our dense RGB-D odometry algorithm \cite{GutierrezGomez2016} which presented the novelty of minimising a geometric error parametrised in inverse depth as it fits better the error model of the depth sensor noise. Relying only on the camera tracking for localisation is prone to produce large deformations in the trajectory and the reconstructed map of the environment as the incremental estimate of the odometry drifts. To correct this we incorporate a place recognition and loop closure module which is based in a Bag of Words scheme using ORB descriptors \cite{MurArtal2014}. For the 3D reconstruction of the environment we use a keyframe-based approach where spatially close frames are integrated in one single keyframe to reduce the depth sensor noise. To build the 3D model the point cloud of each new keyframes are concatenated to the map point cloud, avoiding the addition of 
redundant points. Compared to volumetric approaches, 3D mapping on keyframes is more memory efficient as well as allows to better handle the uncertainty in the depth measures during the frame fusion process. In the experiments we show that our method obtains very accurate results in the trajectory estimation compared against a large selection of the most successful state of the art methods in RGB-D mapping.

Additionally, we present a procedure to completely calibrate RGB-D sensor. The main feature of the proposed method compared to most works in the literature is that in order to calibrate the conversion from the measured depth to the real depth we do not rely in any estimate obtained from the depth camera to generate the ground truth depth. The code has been made available in a public repository for research purposes.

The paper is divided as follows. In Section \ref{sec:rw} we make a revision of the existing method on RGB-D SLAM and 
odometry, as well as RGB-D calibration. Section \ref{sec:rigid_body} is dedicated to explain some preliminary concepts 
of the mathematical representation of rigid body motions. In Section \ref{sec:calibration}, we present the models and 
the pipeline followed for RGB-D camera calibration. Section \ref{sec:warping_functions} presents the different image 
warping approaches that are used in our SLAM approach. Section \ref{sec:dense_alignment} explains our algorithm for 
dense frame alignment required to compute the inter-camera motion and Section \ref{sec:slam_system} describes the 
complete SLAM system with its different modules. In sections \ref{sec:experiments} and \ref{sec:conclussions} we present 
the experiments and the final conclusions.

\section{Related Work} \label{sec:rw}

Maybe the best example of a feature-based RGB-D SLAM system is the approach proposed by Endres \etal \cite{Endres2012}. To estimate the camera motion they first compute an initial seed by RANSAC-based 3D alignment of sparse features, followed by a refinement step where they only minimise the point cloud alignment error from the ICP algorithm. The method is improved in \cite{Endres2014}, including an Environment Measurement Model to prune wrong motion estimates which passed undetected in the RANSAC and ICP steps. The system is able to close loops taking a random sample from a set of keyframes and each frame-to-frame motion calculation between candidates for loop closure is parallelised, thus the computational cost is close to the frame rate of the camera, between 5 and 15 Hz.

KinectFusion by Newcombe \etal \ \cite{KinectFusion} was a ground-breaking contribution in dense RGB-D SLAM. 
KinectFusion is composed by two different modules, one for camera tracking and one for dense volumetric mapping. For 
each new frame first the motion is estimated by frame-to-model ICP alignment of depth maps, \ie, current depth map is 
aligned with the depth map raycasted from a voxelized 3D model. Then, current depth map is integrated in the 3D model 
using a truncated signed distance function.

The main constraint of KinectFusion is its limitation to small workspaces, 
which was nevertheless solved in latter works by using a cyclical buffer to shift the volume as the camera explores the 
environment \cite{Bondarev2013}, \cite{Whelan2013ICRA}. However, artefacts are likely to appear in the reconstructed 3D 
model when an area is revisited. This is solved in recent works by Whelan \etal \cite{Whelan2014,Whelan2016} where a 
loop closure back-end is introduced. This back-end is able to correct potential 
artefacts in the 
3D volume by enforcing the loop constraints not only on the camera trajectory but also on the dense 3D map using a 
deformation graph.

Dong \etal \cite{Dong2014} propose the combination of KinectFusion with a robust odometry algorithm based on keypoint 
matching and RANSAC alignment between frames. This step  provides an 
initial guess of the sensor pose which is further refined by the ICP step of KinectFusion.

Bylow \etal \cite{Bylow2013}  proposed   a   method   which,    as
KinectFusion,  uses  only  the  depth  fusion,  but  rather  than
raycasting a depth map from the model for posterior ICP-
alignment, the camera is tracked by directly minimising the
signed distance function between the current warped depth
map and the model surface defined in the voxelised volume.
This  results  in  a  better  accuracy  and  similar  real-time
computational performance compared to KinectFusion.

Jaimez and Gonz\'alez-Jim\'enez \cite{Jaimez2015} develop a fast visual odometry method using only depth images. 
Their method performs frame-to-frame tracking at a camera frame rate of 60Hz and it can be extended to any kind of 
range sensor.

Following the paradigm of working on 3D models \cite{KinectFusion}, Stuckler and Behnke propose in \cite{Stuckler2012} 
and \cite{Stuckler2014} converting the RGB and depth images into multi-resolution surfel maps by using a voxel octree 
representation. Each surfel maintains a shape-texture descriptor, which guide data association between surfels in 
different maps during camera pose estimation. To alleviate the odometry drift they register the current frame with 
respect to the latest keyframe. A new keyframe is inserted when camera motion w.r.t. last keyframe is large enough. They 
also propose a loop closure technique where loop closure candidates are randomly sampled from a probability density 
function which positively weights the selection of spatially closer keyframes.

Kerl \etal \ \cite{Kerl2013IROS} propose an RGB-D SLAM system where camera motion is estimated with respect to 
keyframes, which are switched following an entropy-based criteria. They include of a simple but effective loop closure 
method based on keyframes spatial proximity to further refine the final odometry estimation. The final 3D is obtained by 
simply joining all the raw depth maps from the keyframes.

Regarding the $3$D reconstruction, though RGB-D sensors already provide a dense depth map of the scene, raw depth maps direct from the sensor show a large noise which grows quadratically with the distance of the observed point. For this reason, in order to obtain an accurate dense 3D reconstruction of the environment one needs to fuse multiple consecutive depth maps. As mentioned earlier in this section approaches like KinectFusion, integrate each of the incoming frames in a 3D volume of voxels. 

Contrary to volume-based approaches, Meilland and Comport \cite{Meilland2013b}, integrate the depth map of each received frame in a set of close keyframes. To track every new frame they use the integrated keyframes to synthesise a reference keyframe from which the new frame is tracked. Though not reporting the use of any loop closure technique their method shows a compelling accuracy in terms of Absolute Trajectory Error, keeping a low trajectory drift.

When comparing both approaches for 3D mapping, volumetric fusion can be viewed as the integration of every incoming frame within a set of concatenated 2D slices, with each slice corresponding to a quantised depth value. This representation in many depth slices can handle occlusions; however it demands large amounts of memory and also, in the process of quantisation and integration of depth measures in the volume, important information about the depth uncertainty is lost. 

In keyframe fusion in turn, integration is performed in a single slice represented by the depth map of the frame at which the keyframe was initialised. The main benefits are two. First, a more lightweight and efficient memory representation and secondly, since integration is performed in the frame of the original image space without discretising the depth values, depth uncertainty can be rigorously handled. The main disadvantage of this representation is that, with a single slice, occlusions between elements in the scene cannot be handled. In our keyframe-based approach, since we generate new keyframes periodically this is not a problem. Also it is worth remarking that a keyframe representation comes on handy when performing applying state of the art appearance-based loop closure strategies in visual SLAM, since it provides a direct mapping of the 3D scene to the RGB image of the keyframe. Furthermore, in our keyframe integration scheme, as in our previous work on direct RGB-D odometry, we take advantage of 
the usage of inverse depth maps instead of depth maps, which allows us to use a constant uncertainty-based tolerance for the integration of a new inverse depth value.

Having a custom calibration for one specific camera is key step to achieve a good precision in SLAM. For this reason another objective of this paper is to obtain a RGB-D SLAM system which can use a custom calibration, as well as providing a calibration pipeline to obtain an accurate calibration. Note that the calibration of an RGB-D sensor is more complex than for RGB ones since it requires first the spatial transformation between the RGB and depth cameras, and secondly a function to compute the depth from the raw disparity maps computed by the camera. In the literature, Smisek \etal \cite{Smisek2011} perform a stereo calibration of RGB and depth cameras, using the infra-red channel of the later. For the correction of a spatial depth distortion they use images of a flat surface, and correct the error wrt. the best fitting plane. Herrera \etal \cite{Herrera2012}, propose a similar approach, but instead they use the depth image for the stereo calibration and perform the distortion correction in the disparity 
space rather than depth. Di Cicco \etal \cite{DiCicco2015} propose a method for depth distortion correction where some images of a wall are taken. Then a plane is fitted to the 
pixels in the centre of the images, under the assumption that these pixels suffer less distortion. Finally a pattern of 
depth multipliers for distortion correction is obtained by regression. Unsupervised approaches for the depth camera 
intrinsic calibration are proposed by Zhou and Koltun \cite{Zhou2014} and Teichman \etal \cite{Teichman2013}, where the 
patterns for depth undistortion are computed taking a ground truth trajectory and map built from a SLAM algorithm. In 
general, the described approaches tend to use the depth images we want to undistort to generate the training examples to 
compute the correction parameters for those distorted images.  In \cite{Zeisl2016}, Zeisl and Pollefeys are aware of 
this issue and propose an auto-calibration method, where a ground truth of camera poses and sparse map points is 
generated by a Structure from Motion algorithm using solely RGB images. In this work we follow the indication in 
\cite{Zeisl2016} of generating the ground truth for depth undistortion from RGB data 
only. However in our case the calibration process is rather supervised by the user. Though it may be a slower and more complex process, we observe that accuracy is increased, since a supervised method allows first to introduce some measure of the scale such as some distance between points, and secondly to control the scene to be only a textured planar surface so that we can generate a dense depth image ground truth from camera poses computed by photogrammetry and the plane constraint. 

%

\section{Preliminaries: Rigid Body Transformations}\label{sec:rigid_body}

This section introduces some concepts of the theory of rigid body transformations which are widely used in this work. For a more complete explanation we refer the reader to \cite{Murray94}. The motion of a rigid body in the 3D space is described by a rotation and a translation with respect to a reference frame. It is usually denoted by a $4\times4$ matrix:

\footnotesize

\begin{align}
 \tensor[_{W}]{\ve{{T}}}{}^{{C}}=\left(
 \begin{array}{cc}
  \tensor[_W]{\ve{{R}}}{}{}^{{C}} & \ve{{t}}_W^C \\[4pt]
  \ve{0}^T & 1
 \end{array}
 \right),
\end{align}

\normalsize


\noindent which reads as the translation and rotation of reference frame \smalleq{$C$}, which is attached to the body, wrt. another reference frame \smalleq{$W$}. \smalleq{$\tensor[_W]{\ve{{R}}}{}{}^{{C}}\in \mathbb{SO}(3)$} and \smalleq{$\ve{{t}}_W^C\in \mathbb{R}^3$} are respectively the rotation matrix and the translation vector.  The rigid body transformations belong then to the product space of \smalleq{$\mathbb{R}^3$} and \smalleq{$\mathbb{SO}(3)$} and is denoted as the Special Euclidean Group, \smalleq{$\mathbb{SE}(3)=\mathbb{R}^3\times \mathbb{SO}(3)$}.

The group \smalleq{$\mathbb{SO}(3)$} induces an over-parametrisation of the rotations by $3$x$3$ matrices with only $3$ 
DoF, which is a source of problems in optimisation or handling rotation uncertainty. To alleviate this it is usual to 
switch between a matrix and an axis-angle representation consisting on a $3$D vector, 
\smalleq{$\bm{\theta}\in\mathbb{R}^3$}. This representation is also denoted as the Lie Algebra 
\smalleq{$\mathfrak{so}(3)$} of \smalleq{$\mathbb{SO}(3)$}. The exponential map $\exp: \mathbb{R}^3 \rightarrow 
\mathbb{SO}(3)$ is used to convert a rotation from axis-angle into a matrix representation:

\footnotesize

\begin{align}
 \exp_{\mathbb{SO}(3)}(\bm{\theta}) = \exp(\mCross{\bm{\theta}}) = \ve{I} + \frac{\sin \norm{\bm{\theta}}}{\norm{\bm{\theta}}}\mCross{\bm{\theta}} + \frac{1-\cos\norm{\bm{\theta}}}{\norm{\bm{\theta}}}\mCross{\bm{\theta}}^2
\end{align}

\normalsize

\noindent where $\mCross{\cdot}$ denotes the skew symmetric $3$x$3$ matrix built from a 3D vector. A rotation matrix is converted to the axis-angle representation through the logarithmic map $\log: \mathbb{SO}(3) \rightarrow \mathbb{R}^3$

\footnotesize

\noindent\begin{tabularx}{\linewidth}{@{}XX@{}}
      \begin{align}
\log_{\mathbb{SO}(3)}(\ve{R}) = \frac{\norm{\bm{\theta}}(\ve{R}-\ve{R}^T)}{2\sin\norm{\bm{\theta}}}  
\end{align}&
\begin{align}
 \norm{\bm{\theta}} = \arccos\left(\frac{\tr(\ve{R})-1}{2}\right)
\end{align}
\end{tabularx}

\normalsize

The notion of exponential and logarithmic map can also be defined in the Group \smalleq{$\mathbb{SE}(3)$}, where an element in its Lie algebra $\mathfrak{se}(3)$ is given by an screw motion \smalleq{$\bm\xi = (\bm{v}, \ \bm{\theta})^T$}, with 6 DoF (3 for rotation \smalleq{$\bm{\theta}$}, and 3 for translation \smalleq{$\bm{v}$}) :

\footnotesize

\begin{align}
 \exp_{\mathbb{SE}(3)}(\bm\xi) = \exp\left(
 \begin{array}{cc}
  \mCross{\bm{\theta}} & \bm{v} \\[4pt]
  \ve{0}^T & 1
 \end{array}
 \right) = 
 \left(
 \begin{array}{cc}
  \ve{R} & \ve{Q}(\bm{\theta})\bm{v} \\[4pt]
  \ve{0}^T & 1
 \end{array}
 \right)
\end{align}

\begin{align}
 \log_{\mathbb{SE}(3)}(\ve{T}) = \left(
 \begin{array}{c}
  \ve{Q}(\bm\theta)^{-1}\ve{t} \\[4pt]
  \log(\ve{R})
 \end{array}
 \right)
\end{align}

\normalsize

\noindent where \smalleq{$\ve{Q} = \int_0^1\exp(\tau\bm{\theta}))d\tau$} and yielding \smalleq{$\ve{t} = \ve{Q}(\bm{\theta})\bm{v}$}. To avoid the rotation having an effect on the translation, in this paper we redefine the exponential and logarithmic maps to act separately in the rotation and translation, that is:

\footnotesize

\begin{align}
 \exp_{\mathbb{SE}(3)}(\bm\xi) = \left(
 \begin{array}{cc}
  \exp\left(\mCross{\bm{\theta}}\right) & \bm{v} \\[4pt]
  \ve{0}^T & 1
 \end{array}
 \right) 
\end{align}

\begin{align}
 \log_{\mathbb{SE}(3)}(\ve{T}) = \left(
 \begin{array}{c}
  \ve{t} \\[4pt]
  \log(\ve{R})
 \end{array}
 \right)
\end{align}

\normalsize

To simplify the notation, from here in advance exponential and logarithmic functions appearing without the group 
sub-index will refer to the map of the defined above which matches with the function arguments. 


\normalsize

\noindent

\section{Camera model and calibration}\label{sec:calibration}

A RGB-D sensor is composed by one RGB and one infra-red (IR) camera, as well as a infra-red emitter which projects a 
dense infra-red 2-dimensional pattern. The projected IR pattern is perceived by the IR camera, and by matching it to a 
reference pattern obtained at a known depth, a dense disparity map can be computed in a similar way as conventional 
stereo systems can do sparsely in parts of the image with high intensity gradient. A RGB-D sensor can provide RGB, IR 
and depth images, where the IR and depth images share the same reference frame and RGB and IR cameras are separated and 
thus the corresponding images have different coordinate systems.  RGB-D cameras include a built-in calibration in their 
hardware which allows for fast registration of the depth image to the RGB frame. However, this factory calibration does 
not account for possible imperfections for a particular camera, and thus a custom calibration is necessary to achieve 
better projection and registration model for a RGB-D sensor.

\subsection{RGB and IR intrinsics}

Both cameras can be modelled by a typical pin-hole camera model with the addition of radial and tangential distortion correction \cite{Brown66}. A scene point \smalleq{$\ve{X}=\left(X,\ Y,\ Z\right)^T$} is first projected on a normalised plane obtaining the undistorted coordinates \smalleq{$\ve{m}_{u} = \left(X/Z, \ Y/Z, \ 1\right)^T$}. Then the distortion correction is applied:

\smalleq{\begin{align}
 \ve{m}_{d} = \ve{h}(\ve{m}_u)  
 & = (1+k_1r_u^2+k_2r_u^4+k_5r_u^6)\ve{m}_{u} + \notag \\ 
 &+\left(
 \begin{array}{c}
  2k_3{m}_{u,x}{m}_{u,y} + k_4(r_u^2+2{m}_{u,x}^2) \\
  2k_4{m}_{u,x}{m}_{u,y} + k_3(r_u^2+2{m}_{u,y}^2) \\
  -(k_1r_u^2+k_2r_u^4+k_5r_u^6)
 \end{array}
\right) \label{eq:proj_dist}
\end{align}}

\noindent where \smalleq{$r_u = \sqrt{m_{u,x}^2 + m_{u,y}^2}$}. The vector \smalleq{$\ve{k} = \left(k_1, \ k_2, \ k_3, \ k_4,\ k_5\right)^T$} encapsulates the distortion parameters. This step is followed by a linear projection from the normalised plane in the final image plane:

\smalleq{\begin{align}
 \ve{p} = \underbrace{\left[
 \begin{array}{ccc}
  f_x & 0 & c_x \\
  0 & f_y & c_y \\
  0 & 0 & 1
 \end{array}
 \right]}_{\ve{K}}
  \ve{m}_d  \label{eq:proj_linear}
\end{align}
}

\noindent where \smalleq{$\ve{K}$} is the pinhole model calibration matrix including the focal length \smalleq{$\ve{f} = \left(f_x,\ f_y\right)^T$} and the principal point \smalleq{$\ve{c} = \left(c_x,\ c_y, \ 1 \right)^T$}. All the previous steps can be encapsulated in a projectivity function \smalleq{$\ve{p} = \Scale{1.3}{\bm{\pi}}(\ve{X})$}. Image points can be also lifted to the 3D world with the inverse model:

\smalleq{\begin{align}
 \ve{X} = \Scale{1.3}{\bm{\pi}}^{-1}(\ve{p}) = \ve{h}^{-1}(\ve{K}^{-1}\ve{p}).
\end{align}}

Note that in this last equation, for \smalleq{$\ve{X}$} arbitrarily \smalleq{$Z=1$} since depth cannot be recovered from 
one single projection. It is necessary to know the depth \smalleq{$\mathcal{Z}(\ve{p})$} in a pixel to scale 
\smalleq{$\ve{X}$} and obtain a full 3D point.

The depth camera provides a depth image which is obtained from the disparity map using a generic conversion function. Thus, additional calibration parameters are needed to improve the depth map accuracy for a specific camera. We apply the following model to the inverted raw depth map \smalleq{$\mathcal{W}_m = \frac{1}{\mathcal{Z}_m}$} measured by the camera :


\smalleq{
\begin{align}
\mathcal{W}_d(\ve{p}) &= \beta_1\mathcal{W}_m(\ve{p}-\ve{p}_0) + \beta_0  \label{eq:depth_linear_correction}
\end{align}
}

\noindent where \smalleq{$\beta_0$} and \smalleq{$\beta_1$} are the calibration parameters, and \smalleq{$\ve{p}_0 = \left(4,\ 4, \ 0\right)^T$} is a pixel shift caused by the correlation window used internally by the depth sensor to compute the disparity \cite{KinectCalibration}. The purpose of this function is to correct the the disparity-to-depth conversion computed internally by the sensor using factory constants.

However, the depth sensor introduces also a spatially dependent error. To correct this distortion we apply an undistortion function as follows:

\smalleq{
\begin{align}
 \mathcal{W}_u(\ve{p}) &= \mathcal{D}_1(\ve{p})\mathcal{W}_d(\ve{p}) + \mathcal{D}_0(\ve{p}), \label{eq:depth_spatial_correction}
\end{align}}

\noindent where $\mathcal{D}_0(\ve{p})$ and $\mathcal{D}_1(\ve{p})$ are spatially varying offset and multiplying factor respectively. Each of these terms  is calibrated with $8$ parameters which are the coefficients of a polynomial in the 2D image coordinates:


\smalleq{\begin{align}
 \mathcal{D}_i(\ve{p}) &= q_{i,0} + q_{i,1}r^2+q_{i,2}r^4+q_{i,3}r^6+q_{i,4}m_x+q_{i,5}m_y+\notag \\
 &+q_{i,6}m_xm_y+q_{i,7}m_x^2m_y+q_{i,8}m_xm_y^2,
\end{align}}

\noindent where \smalleq{$\ve{m} = \ve{K}^{-1}\ve{p}$} and \smalleq{$r = \sqrt{m_x^2+m_y^2}$}.

\subsection{RGB and depth cameras extrinsics}

The relative pose between the color and the depth camera frames is denoted by the spatial transformation matrix \smalleq{$\tensor[_{D}]{\ve{{T}}}{}^{{C}}=\left\lbrace\tensor[_D]{\ve{{R}}}{}{}^{{C}}, \ve{{t}}_D^C\right\rbrace$}. It is required to register the inverse depth image in the color camera frame.

\subsection{Calibration pipeline}

The calibration of the RGB-D camera is performed in two steps. The first step consists in a conventional calibration of the IR and RGB cameras intrinsics and stereo transformation using the Bouguet calibration Toolbox \cite{BouguetCalibration}. With this procedure we obtain the RGB and depth intrinsics \smalleq{$\ve{f}_{\left\lbrace C,D\right\rbrace}$, $\ve{c}_{\left\lbrace C,D\right\rbrace}$, $\ve{k}_{\left\lbrace C,D\right\rbrace}$} and the spatial transformation \smalleq{$\tensor[_{D}]{\ve{{T}}}{}^{{C}}$} between the RGB and depth camera frames.


The second step in the calibration is the computation of the depth intrinsics for the correction of the raw inverse 
depth maps. For this step, we need $M$ synchronised depth and RGB shots taken at different distances from a wall filling 
the whole image with a checker-board pattern and some posters attached to it. The real inverse depth map is computed by 
photogrammetry from the RGB images. After corner extraction, image matching and applying bundle adjustment techniques we 
obtain the pose \smalleq{$\tensor[_{W}]{\ve{{T}}}{}^{{C}}_i$} of each of the $M$ RGB frames and a 3D sparse point cloud. 
Since the dimensions of the checker-board pattern are known, the estimated variables are in the true scale.  The plane 
\smalleq{$\bm\Pi = (\ve{n}_W, d)$} which best fits the point cloud is also computed. 

With the poses \smalleq{$\tensor[_{W}]{\ve{{T}}}{}^{{C}}_i$}, and the stereo camera intrinsics from the previous step all the pixels in the inverse depth image can be lifted and represented in the global reference frame:

\footnotesize

{\begin{align}
 \ve{X}_W^{i}(\ve{p})&=\tensor[_{W}]{\ve{{R}}}{}^{{C}}_i\tensor[_{C}]{\ve{{R}}}{}^{{D}}\frac{1}{\mathcal{W}(\ve{p})}\Scale{1.3}{\bm{\pi}}^{-1}_D(\ve{p})+\tensor[_{W}]{\ve{{R}}}{}^{{C}}_i\ve{t}_C^D+\ve{t}_{W,i}^C \notag \\
 &=\tensor[_{W}]{\ve{{R}}}{}^{{D}}_i\frac{1}{\mathcal{W}(\ve{p})}\Scale{1.3}{\bm{\pi}}^{-1}_D(\ve{p})+\ve{t}_{W,i}^D, 
\end{align}}

\normalsize

And by applying the restriction of all the image points being contained in a plane \smalleq{$\ve{n}_W^T\ve{X}_W^{i}(\ve{p})+ d = 0$},  we can compute a ground truth for the inverse depth images:

\footnotesize

\smalleq{\begin{align}
  \mathcal{W}_{GT,i}(\ve{p}) = - \frac{\ve{n}_W^T\tensor[_{W}]{\ve{{R}}}{}^{{D}}_i\Scale{1.3}{\bm{\pi}}^{-1}_D(\ve{p})}{\ve{n}_W^T\ve{t}_{W,i}^D+d} \label{eq:plane_restriction}
\end{align}}

\normalsize

Then, the intrinsic depth correction parameters are computed in two steps. In the first step we compute only the parameters of the linear relation:
%

\footnotesize

\begin{align}
 \operatorname*{argmin}_{\bm{\beta}} \sum_{i=1}^M\sum_{\substack{\ve{p}\in\Omega \\ \ve{p}-\ve{p}_0\in\Omega}}\norm{\mathcal{W}_{GT,i}(\ve{p}) - (\beta_1\mathcal{W}_{m,i}(\ve{p}-\ve{p}_0)+\beta_0)}^2
\end{align}

\normalsize

In the second step after applying \eqref{eq:depth_linear_correction} we compute the coefficients of the distortion polynomials:

\footnotesize

\begin{align}
 \operatorname*{argmin}_{\bm{q}_0, \bm{q}_1} \sum_{i=1}^M\sum_{\substack{\ve{p}\in\Omega}}\norm{\mathcal{W}_{GT,i}(\ve{p}) - (\mathcal{D}_1(\ve{p})\mathcal{W}_{d,i}(\ve{p})+\mathcal{D}_0(\ve{p}))}^2
\end{align}

\normalsize

The reason of optimising in two separate steps is obtaining a calibration where we can choose whether use or not use the 
spatial distortion correction. 

\section{Warping functions}\label{sec:warping_functions}

A key element in direct methods for visual slam is the image warping functions. These are need for distortion correction or to obtain representations of the scene in some image as viewed from another different pose. Depending on what we need to achieve we use different warping approaches. 

\subsection{Inverse warping}

Given a source image \smalleq{$\mathcal{I}_A$} and a destination image \smalleq{$\mathcal{I}_B$}, inverse warping 
consists into mapping every pixel \smalleq{$\ve{p}\in\Omega$} in the destination image to a position in the source 
image, and then obtain the corresponding value by bilinear interpolation at the source:

\smalleq{\begin{align}
 \mathcal{I}_B(\ve{p}) = \mathcal{I}_A(\ve{f}_w(\ve{p}))
\end{align}}

\noindent where the \smalleq{$\ve{f}_w$} is the warping function. We use inverse warping for undistortion of intensity and inverse depth maps, where \smalleq{$\ve{f}_w(\ve{p}) = \ve{K}\ve{h}(\ve{K}^{-1}\ve{p})$}.

%
%
%

\subsection{Forward registration}

Forward registration is required when we need to represent an inverse depth map in another reference frame, like for 
example the one of the RGB camera. Generically, given the spatial transformation 
\smalleq{$\tensor[_{A}]{\ve{{{T}}}}{}^{{B}}$} and assuming that images have been undistorted, a pixel \smalleq{$\ve{p}$} 
in a depth map \smalleq{$\mathcal{W}_A$} at source frame $A$ is mapped to a pixel \smalleq{$\ve{p}_B$} in a destination 
frame $B$ through the following equation:

\smalleq{\begin{align}
 \frac{1}{w_B}\ve{p}_B = \tensor[_{B}]{\ve{{\tilde{R}}}}{}^{{A}}\left(\frac{1}{\mathcal{W}_A(\ve{p})}\ve{p}-\ve{\tilde{t}}_A^B\right), \label{eq:3d_registration}
\end{align}}

\noindent where \smalleq{$\tensor[_{B}]{\ve{{\tilde{R}}}}{}^{{A}} = \ve{K}_B\tensor[_{B}]{\ve{{{R}}}}{}^{{A}}\ve{K}_A^{-1}$} and \smalleq{$\ve{\tilde{t}}_A^B = \ve{K}_A\ve{{t}}_A^B$}.

A naive approach would be to compute \smalleq{$\ve{p}_B$}, round to the nearest pixel location and then update the 
inverse depth value at this location. Unfortunately this approach is liable to produce holes in the new registered depth 
map. Instead we use a solution based on the relief texture mapping algorithm \cite{Oliveira2000}, performing the 
registration in two steps.

In the first step the depth image is registered to an intermediate frame $B_\ve{t}$ applying only the shift due to translation:

\smalleq{\begin{align}
 \frac{1}{w_{B_\ve{t}}}\ve{p}_{B_\ve{t}} = \frac{1}{\mathcal{W}_A(\ve{p})}\ve{p}-\ve{\tilde{t}}_A^B. \label{eq:register_trans}
\end{align}}

From this equation we obtain:

\footnotesize

\noindent\begin{tabularx}{\linewidth}{@{}XX@{}}
      \begin{align}
w_{B_\ve{t}} = \frac{\mathcal{W}_A(\ve{p})}{1-\mathcal{W}_A(\ve{p})\ve{e}_\ve{z}^T\ve{\tilde{t}}_A^B}
\end{align}
&
\begin{align}
\ve{p}_{B_\ve{t}} = \frac{\ve{p}-\mathcal{W}_A(\ve{p})\ve{\tilde{t}}_A^B}{1-\mathcal{W}_A(\ve{p})\ve{e}_\ve{z}^T\ve{\tilde{t}}_A^B}
\end{align}
\end{tabularx}

\normalsize

\noindent to prevent the arising of holes we map the whole pixel surface to the frame ${B_\ve{t}}$. Adding \smalleq{$\bm{\delta}\ve{p} = (\delta{p_x}, \ \delta{p_y}, \ 0)$} to \smalleq{$\ve{p}$} in  \eqref{eq:register_trans} and assuming that   \smalleq{$\mathcal{W}_A(\ve{p}) = \mathcal{W}_A(\ve{p}+\bm{\delta}\ve{p})$}  \smalleq{$\forall \norm{\bm{\delta}\ve{p}}_\infty \leq 0.5$}, we get:

\smalleq{\begin{align}
 \bm{\delta}\ve{p}_{B_\ve{t}} = \frac{w_{B_\ve{t}}}{\mathcal{W}_A(\ve{p})}\bm{\delta}\ve{p}.
\end{align}}

\noindent Taking the four pixel corners at frame $B$ at \smalleq{$\delta p_x\pm0.5$} and \smalleq{$\delta p_y\pm0.5$} and rounding the corresponding \smalleq{$\ve{p}_{B_\ve{t}}+\bm{\delta}\ve{p}_{B_\ve{t}}$} to the nearest integer, we obtain a square containing all the pixels where \smalleq{$w_{B_\ve{t}}$} should be mapped at frame ${B_\ve{t}}$. For every pixel \smalleq{$\ve{p}_s$} in this square \smalleq{$\mathcal{W}_{B_\ve{t}}(\ve{p}_s)$} is updated if \smalleq{$w_{B_\ve{t}}$} is the greatest or the first inverse depth value computed  in that pixel.

In the second step, the remaining rotation is applied to the warped frame. Noting that the \smalleq{$\tensor[_{B}]{\ve{{\tilde{R}}}}{}^{{A}}$} represents an homography, and thus a linear mapping, inverse warping can be applied taking pixels in the destination frame $B$:

\smalleq{\begin{align}
 \frac{1}{\mathcal{W}_{B}(\ve{p})}\ve{p} = \tensor[_{B}]{\ve{{\tilde{R}}}}{}^{{A}}\frac{1}{\mathcal{W}_{B_\ve{t}}(\ve{p}_{B_\ve{t}})}\ve{p}_{B_\ve{t}}
\end{align}}

From this equation \smalleq{$\ve{p}_{B_\ve{t}} = \tfrac{\tensor[_{A}]{\ve{{\tilde{R}}}}{}^{{B}}\ve{p}}{\ve{e}_\ve{z}^T\tensor[_{A}]{\ve{{\tilde{R}}}}{}^{{B}}\ve{p}}$}, with \footnotesize $\tensor[_{A}]{\ve{{\tilde{R}}}}{}^{{B}} = \left(\tensor[_{B}]{\ve{{\tilde{R}}}}{}^{{A}}\right)^{-1}$\normalsize. Then \smalleq{$\mathcal{W}_{B_\ve{t}}(\ve{p_{B_\ve{t}}})$} is obtained by nearest neighbour interpolation and finally the inverse depth value \smalleq{$\mathcal{W}_{B}(\ve{p}) = \mathcal{W}_{B_\ve{t}}(\ve{p_{B_\ve{t}}})\ve{e}_\ve{z}^T\tensor[_{A}]{\ve{{\tilde{R}}}}{}^{{B}}\ve{p} $} in the registered image.

%

\subsection{Inverse geometric warping}\label{sec:warping2}

If inverse depth and intensity maps are available at both frames $A$ and $B$, inverse warping can be performed by using 
\eqref{eq:3d_registration} as the warping function. Note that since we are performing inverse warping, the roles of the 
frames $A$ and $B$ in this equation is interchanged ($A$ is the destination, and $B$ is the source). This is 
computationally faster than forward warping and is useful, for example, to fuse inverse depth maps taken at different 
poses. The drawback is that occlusions cannot be handled and thus the motion between frames must be small. Given that 
the pixel position \smalleq{$\ve{p}_B$} computed by \eqref{eq:3d_registration}, the images at frame $B*$ warped towards 
frame $A$ are computed by the following relations:

\smalleq{\begin{align}
    \mathcal{I}_{B*}(\ve{p}) &= \mathcal{I}_{B}(\ve{p}_{B}) \\[4pt]
    \mathcal{W}_{B*}\left(\ve{p} \right) &= \frac{\ve{e}_\ve{z}^T\tensor[_B]{\ve{\tilde{R}}}{}^{{A}}\ve{p}}{1-\mathcal{W}_B\left(\ve{p}_B \right)\ve{e}_\ve{z}^T\ve{\tilde{t}}_B^A}\mathcal{W}_B\left(\ve{p}_B \right) 
  \end{align}}

\subsection{GPU implementation}

All the warping functions have been implemented for CUDA capable NVIDIA GPUs. In the inverse warping functions, the 
mapping of source images to the GPU texture memory allows for efficient interpolation via texture fetching. In the case 
of the translation-only warping step in the forward registration procedure, a pixel in the destination images might be 
accessed at the same time by two or more GPU threads for a writing operation. In order to avoid race conditions and 
serialise the access for writing we use atomic operations.

\section{Dense motion estimation}\label{sec:dense_alignment}

Dense frame alignment consists into estimating the camera motion \smalleq{$\tensor[_{A}]{\ve{\hat{T}}}{}^{{B}}=(\tensor[_A]{\ve{\hat{R}}}{}{}^{{B}}, \ve{\hat{t}}_A^B)$} between two frames $A$ and $B$ by pixel-wise minimisation of the photometric and inverse depth error between both frames.


\subsection{Photometric and geometric constraints}

Let us denote two camera frames as \small {$A$} \normalsize  and \small {$B$}\normalsize, at instants \smalleq{$t$} and \smalleq{$t +
\Delta{t}$} respectively.  Given the intensity images
\smalleq{$\mathcal{I}_A$} and \smalleq{$\mathcal{I}_B$}, and
inverse depth maps  \smalleq{$\mathcal{W}_A$}  and  \smalleq{$\mathcal{W}_B$} defined over the image domain
\smalleq{$\Omega\subset\mathbb{P}^2$}, for an image point \small {$\ve{p} = (u \ v \ 1)^T\in\Omega$} \normalsize  in frame \small
{$A$}\normalsize, the following constraints hold:

\small

\begin{align}
 \mathcal{I}_B(\ve{p}+\ve{{\Delta}p}) &=  \mathcal{I}_A(\ve{p})  \label{eq:intCons}\\
 \mathcal{W}_B(\ve{p}+\ve{{\Delta}p}) &= \frac{\mathcal{W}_A(\ve{p})}{1+\mathcal{W}_A(\ve{p})\ve{e}_\ve{z}^T\ve{{\Delta}X_p}}, \label{eq:depthCons}
\end{align}

\normalsize

\noindent where \smalleq{$\ve{{\Delta}X}_{\ve{p}}$} and
\smalleq{$\ve{{\Delta}p}$} are respectively the scene and the optical flow of one point
from frame \smalleq{$A$} to \smalleq{$B$}, and \smalleq{$\ve{e}_\ve{z}^T =
\left( 0 \ \ 0 \ \ 1 \right)$}.
The constraint in intensity assumes constant illumination of one scene point. The second constraint models the change of one point's inverse depth due to the relative motion along the optical axis of the camera.

It can be verified that the scene and the optical flow satisfy the following equation:

\small

\begin{align}
 \mathcal{W}_A(\ve{p})\left(\ve{K}-\ve{p}\ve{e}_\ve{z}^T\right)\ve{{\Delta}X}_{\ve{p}} = (1+\mathcal{W}_A(\ve{p})\ve{e}_\ve{z}^T\ve{{\Delta}X}_{\ve{p}})\ve{\Delta{p}}
\end{align}

\normalsize

With this relation and computing the first order Taylor expansion with respect to the scene flow we get the linearised constraints:

\small

\begin{align}
 \mathcal{W}_A(\ve{p}){\nabla}\mathcal{I}_A(\ve{p})\left(\ve{K}- \ve{p}\ve{e}_\ve{z}^T\right)\ve{{\Delta}X}_{\ve{p}}  +
\mathcal{I}_B(\ve{p}) - {\mathcal{I}_A(\ve{p})}  &=  0 \label{eq:3dConsInt} \\
 \mathcal{W}_A(\ve{p})\left({\nabla}\mathcal{W}_A(\ve{p})\left(\ve{K}-
\ve{p}\ve{e}_\ve{z}^T\right)+\mathcal{W}_B(\ve{p})\ve{e}_\ve{z}^T\right)\ve{{\Delta}X}_{\ve{p}} & + \notag \\+
\mathcal{W}_B(\ve{p}) - \mathcal{W}_A(\ve{p}) &= 0. \label{eq:3dConsDepth}
\end{align}

\normalsize

Since we are estimating the motion between frames we are working under the assumption of a rigid scene.  Assuming a small motion described by the rotation and translation
pair \smalleq{$(\tensor[_A]{\ve{R}}{}{}^{{B}}, \ve{t}_B^A)\in\mathbb{SE}(3)$}  we have:

\small

\begin{align}
\ve{{\Delta}X}_{\ve{p}} &= \left(\tensor[_B]{\ve{R}}{^A}-\ve{I}\right)\ve{X}_A + \ve{t}_B^A \notag \\ 
&= \mCross{\bm{\theta}_B^A}\ve{X}_A +  \ve{t}_B^A +
\mathcal{O}\left(\norm{\mCross{\bm{\theta}_B^A}^2\ve{X}_A}\right) \notag \\
&\approx \ve{t}_B^A - \mCross{\ve{X}_A}{\bm{\theta}_B^A}= \ve{M}(\ve{p})\bm{\xi}_B^A. \label{eq:DeltaX}
\end{align}

\noindent where \smalleq{$\ve{X}_A = \ve{K}^{-1}\ve{p}\frac{1}{\mathcal{W}_A(\ve{p})}$} and \smalleq{$\bm{\xi}_B^A = (\ve{t}_B^A, \
\bm{\theta}_B^A)^T$}. $\mCross{\cdot}$ denotes the antisymmetric matrix from a vector and \smalleq{$\bm{\theta}_B^A$} is the axis-angle representation of \smalleq{$\tensor[_B]{\ve{R}}{}{}^{{A}}$}.


%

\subsection{Iterative optimisation}

\normalsize

The inter-frame motion is computed iteratively in a coarse-to-fine manner using image pyramids, performing a given 
number of iterations at each pyramid level. At the start and every time we step down in the image pyramid, we downsample 
\smalleq{$\left\lbrace\mathcal{I}_{A},\
\mathcal{W}_{A}\right\rbrace$} and compute the gradients \smalleq{$\left\lbrace\nabla\mathcal{I}_{A},\ \nabla\mathcal{W}_{A}\right\rbrace$} at current pyramid level. The initial estimate of the rigid motion is set to the identity, \smalleq{$\tensor[_{A}]{\ve{\hat{T}}}{}^{B}_{(0)} = \ve{I}$}, unless an initial guess is provided. 


At each iteration \smalleq{$\gamma$}, first intensity and depth maps in frame $B$,
\smalleq{$\left\lbrace\mathcal{I}_{B},\
\mathcal{W}_{B}\right\rbrace$}, are warped towards frame $A$
taking the up-to-date motion estimate \smalleq{$\tensor[_{A}]{\ve{\hat{T}}}{}^{B}_{(\gamma)}$}. By using the reverse warping approach described in Section \ref{sec:warping2} we obtain the warped images
\smalleq{$\left\lbrace\mathcal{I}\right._{B}^{(\gamma)},\ \mathcal{W}_{B}^{(\gamma)}\left.\!\right\rbrace$}. These images are downsampled to the pyramid level at which current iteration is taking place. The formulae for the residuals at current iteration are given by:

\footnotesize
\begin{align}
r_\mathcal{I}(\ve{p},\bm{\xi}) &=  \mathcal{W}_A(\ve{p}){\nabla}\mathcal{I}_A(\ve{p}){(\ve{K}{-}
\ve{p}\ve{e}_\ve{z}^T)}\ve{M}(\ve{p})\bm{\xi}+ \notag \\ & \ \ \ \ + 
\mathcal{I}_B^{(\gamma)}(\ve{p}) - {\mathcal{I}_A(\ve{p})} \label{eq:res_int} \\
\notag \\
r_\mathcal{W}(\ve{p},\bm{\xi}) &= \mathcal{W}_A(\ve{p}){\left({\nabla}\mathcal{W}_A(\ve{p}){(\ve{K}{-}
\ve{p}\ve{e}_\ve{z}^T)}{+}\mathcal{W}_B^{(\gamma)}(\ve{p})\ve{e}_\ve{z}^T\right)}\ve{M}(\ve{p})\bm{\xi}+  \notag \\ & \ \ \ \  
+ \mathcal{W}_B^{(\gamma)}(\ve{p}) - \mathcal{W}_A(\ve{p}), \label{eq:res_depth}
\end{align}

\normalsize

The next step is the computation of the update of the motion estimate between frames. This is achieved by optimisation with Iteratively Reweighted
Least Squares algorithm (IRLS) \cite{IRLS}, which results in a linear least-squares problem to be solved at each iteration:

\footnotesize

\begin{align}
{\bm{\hat{\xi}}_{B^{(\gamma)}}^A} = \operatorname*{argmin}_{\bm{\xi}}
\sum_{{\ve{p}}\in\Omega}&\omega{\left(\frac{r_{\mathcal{I}}(\ve{p},\bm{0})-\mu_{\mathcal{I}}}{\sigma_{\mathcal{I}}}\right)}{\frac{r_\mathcal{I}^2
(\ve{p},\bm{\xi})}{\sigma_{\mathcal{I}}^2}} \notag \\
&+\lambda_{\ve{n}}(\ve{p})\omega{\left(\frac{r_{\mathcal{W}}(\ve{p},\bm{0})-\mu_{\mathcal{W}}}{\sigma_{\mathcal{W}}}\right)}{\frac{r_\mathcal{W}^2
(\ve{p},\bm{\xi})}{\sigma_{\mathcal{W}}^2}}, \label{2eq:minimisationIRLS}
\end{align}

\normalsize

The weighting function \smalleq{$\omega(x)$} depends on the assumed probability distribution of the residuals. \smalleq{$\mu_{\left\lbrace\mathcal{I},\mathcal{W}\right\rbrace}$} and \smalleq{$\sigma_{\left\lbrace\mathcal{I},\mathcal{W}\right\rbrace}$} are the location and scaling parameters which
capture the bias and the uncertainty in intensity and inverse depth residuals, and allow for normalisation of residuals 
in different magnitudes. The details on the computation of the weighting function and the location and scaling 
parameters are presented in the next subsection \ref{sec:errordistr_parameters}. To account for noisier depth 
measurements from surfaces more oblique to the camera, \smalleq{$\lambda_{\ve{n}}(\ve{p}) = 
\ve{n}^T\tfrac{\ve{p}}{\norm{\ve{p}}}$} is used to down-weight the inverse depth residuals, where the normal 
\smalleq{$\ve{n}^T$} is computed as:

\footnotesize

\begin{align}
 \ve{n}^T = \alpha\left(\frac{1}{\mathcal{W}_A(\ve{p})}{\nabla}\mathcal{W}_A(\ve{p})\left(\ve{K}-\ve{p}\ve{e}_\ve{z}^T\right)+\ve{e}_\ve{z}^T\right),
\end{align}

\normalsize

\noindent where \smalleq{$\alpha$} is the normalising constant.

The minimisation of \eqref{2eq:minimisationIRLS} leads us to solve the following linear system:

\small

\begin{align}
 \ve{H}^{(\gamma)}\bm{\hat{\xi}}_{B^{(\gamma)}}^A=\ve{b}^{(\gamma)},
\end{align}

\normalsize

and after computing \smalleq{$\bm{\xi}_{B^{(\gamma)}}^A$} we update the inter-frame motion estimate for the next 
iteration:

\small

\begin{align}
 \tensor[_{A}]{\ve{\hat{T}}}{}^{B}_{(\gamma+1)} = 
\inv{
\left(
\begin{array}{cc}
 \exp(\mCross{\bm{\hat{\theta}}_{B^{(\gamma)}}^A}) & \ve{t}_{B^{(\gamma)}}^A \\
 0 & 1
\end{array}
 \right)
}\tensor[_{A}]{\ve{\hat{T}}}{}^{B}_{(\gamma)}.
\end{align}

\normalsize

The covariance of the motion estimate is computed as the inverse of the Hessian. The Hessian depends on the intensity and depth maps gradients computed in the frame $A$. Due to sensor noise it is possible that some pixels provide misleading information making the Hessian larger, which would yield a wrongly optimistic covariance matrix. In order to prevent this, once we have computed the final iteration for the motion estimate, we estimate the Hessian by performing an extra iteration where the motion estimate is not updated and where \smalleq{$\left\lbrace\mathcal{I}_{A},\ \mathcal{W}_{A}\right\rbrace$} are applied a bilateral filter prior to the computation of their gradients:

\small 

\begin{align}
  \bm{\Sigma}_{A}^{A\rightarrow{B}} = \bm{\Sigma}_{A}^{B\rightarrow{A}} = \ve{H}_{filt}^{-1}
\end{align}

\normalsize

\subsection{Error distribution and robust optimisation}\label{sec:errordistr_parameters}


The underlying assumption behind the IRLS method is that the residuals we want to minimise are modelled by a probability 
density function (pdf). This function is related to the weights $\omega(x)$ we apply in the cost function. The purpose 
of IRLS is that, by selecting a pdf with heavier tails than the typically assumed Gaussian distribution, gross errors 
can be accommodated and and thus we gain robustness to outliers. In this work we use the student t-distribution whose 
pdf is given by:

{\begin{align}
 f_{St}(r_i\mid\mu,\ \sigma, \ \nu) = \frac{\Gamma\left(\tfrac{\nu+1}{2}\right)}{\sqrt{\nu\pi}\Gamma\left(\tfrac{\nu}{2}\right)\sigma}\left(1+\frac{x_i^2}{\nu}\right)^{-\tfrac{\nu+1}{2}} \label{eq:Student_pdf}
\end{align}}

\noindent with $x_i = \frac{r_i-\mu}{\sigma}$, and where $\mu$, $\sigma$ are the location, and the scale of the pdf, 
while $\nu$ is a characteristic parameter of the Student's t-distribution and denotes the degrees of freedom. In robust 
estimation the parameter $\nu$ is associated to the achieved robustness. Taking the logarithm of \eqref{eq:Student_pdf}  
and eliminating constants we obtain the log-likelihood function:

\smalleq{\begin{align}
 \log\mathcal{L}(\mu,\ \sigma, \ \nu \mid r_i) = &\ln\left(\Gamma\left(\tfrac{\nu+1}{2}\right)\right) - \ln\left(\Gamma\left(\tfrac{\nu}{2}\right)\right) \notag \\
 &-\ln(\sigma) + \tfrac{\nu}{2}\ln(\nu) - \underbrace{\tfrac{\nu+1}{2}\ln(\nu+x_i^2)}_{\rho(x_i,\ \nu)} \label{eq:logLikelihood_St}
\end{align}}

\normalsize

The location and scaling parameters of the residuals are obtained by maximisation of their joint log-likelihood function. That is:

\smalleq{\begin{align}
 \left[\mu, \ \sigma\right] &= \operatorname*{argmax}_{\mu, \sigma} \sum_i \log\mathcal{L}(\mu,\ \sigma, \ \nu \mid r_i) \notag \\
  &= \operatorname*{argmin}_{\mu, \sigma} \sum_i \ln(\sigma)+\rho(x_i, \nu)
\end{align}}

Taking the derivatives with respect to $\mu$ and $\sigma$ we obtain respectively the following equations:

\smalleq{\begin{align}
 \sum_i \omega(x_i,\ \nu)({r_i-\mu}) = 0 \\
 \sum_i \left(1-\omega(x_i,\ \nu)\frac{(r_i-\mu)^2}{\sigma^2}\right) = 0 
\end{align}}

\noindent where \smalleq{$\omega(x,\ \nu) = \frac{1}{x}\frac{\partial{\rho(x,\ \nu)}}{\partial{x}}$}. This system of equations can be solved iteratively using the previous estimates of \smalleq{$\mu$} and \smalleq{$\sigma$} to compute \smalleq{$\omega(x_i,\ \nu)$} before each iteration.

Note that up to this point we have ignored the estimation of the parameter $\nu$. Typically this parameter is set to 
\smalleq{$\nu = 5$} which provides enough robustness and also behaves well in case the residuals have a Gaussian 
distribution. However, \smalleq{$\nu$} can be estimated as well, which allows to adapt the required robustness to the 
distribution of current residuals. By maximising \eqref{eq:logLikelihood_St} with respect to \smalleq{$\nu$} 
\cite{estimationNuStudent} we obtain the following non-linear equation:

\small

\begin{align}
 \sum_{i}\left(-\phi\left(\frac{\nu}{2}\right) + \ln\left(\frac{\nu}{2}\right)  + \phi\left(\frac{\nu+1}{2}\right) - \ln\left(\frac{\nu+1}{2}\right) \right. \notag \\
 \left. \phantom{\ln\left(\frac{\nu+1}{2}\right)} + 1 + \ln(\omega(x_i,\nu))-\omega(x_i,\nu)\right) = 0, \label{eq:robustnessValue}
\end{align}

\normalsize

\noindent where \smalleq{$\phi(y)$} denotes the digamma function. Note that ideally robustness parameters should be estimated at the same time as location and scaling. However, unlike location and scaling equations, fixing the weights in \eqref{eq:robustnessValue} with the previous estimate does not produce a linear system, requiring from some sub-iterations to compute $\nu$ within current iteration. Moreover fixing \smalleq{ $\nu$} in the weights to its previous value produces a slow convergence.

However letting \smalleq{$\nu$} to be an unknown in the weights implies to perform  a sub-loop of GPU reduction 
operations within the loop in which the distribution parameters are estimated. Therefore, to keep computational cost low 
we estimate first the final location and scaling values for intensity and depth residuals fixing \smalleq{$\nu=5$}, and 
then we solve \eqref{eq:robustnessValue} with the computed values using the bisection method assuming 
\smalleq{$\nu\in[2,10]$} with a lower \smalleq{$\nu$} implying more robustness to outliers.  This is performed both for 
the intensity and inverse depth residuals yielding \smalleq{$\nu_\mathcal{I}$} and \smalleq{$\nu_\mathcal{W}$},

In dense RGB-D tracking the main source of outliers are the high inverse depth residuals caused by occlusions which are due to dynamic objects or areas which simply become revealed or hidden  by the camera motion, \ie, outliers are caused by large inverse depth residuals rather than intensity ones. For this reason and because too much robustness can lead to a bad or slow convergence to the solution, we select the highest of the computed \smalleq{$\nu_\mathcal{W}$} and \smalleq{$\nu_\mathcal{I}$} for the intensity residuals.


\smalleq{
\begin{align}
 \nu_\mathcal{I} = \max(\nu_\mathcal{I}, \nu_\mathcal{W}).
\end{align}}

\normalsize

A naive computation of the location, scaling and robustness parameters would use all the residuals. Given that every residual in intensity and inverse depth corresponds to a single pixel, the number of samples would be extremely large, more than \smalleq{$300000$} samples at the highest resolution level, which would involve a high computational cost. To reduce this burden we compute the scaling parameters in a sample with a maximum size of \smalleq{$N=19200$} points obtained by systematic selection. This sample size guarantees first an integer stride for pixel sampling at all the considered resolutions; and secondly, after leading an statistical analysis, it also guarantees a relative precision of \smalleq{$0.03$} with a confidence level of \smalleq{$99.7\%$}. 

\subsection{Dense frame covisibility ratio}\label{sec:covisibility_ratio}

Computing the covisibility score between two given frames is a key element to prune redundant frames, selection of reference frames or keyframes, or computing visibility graphs. In this work, taking advantage of the availability of dense depth maps, we use a covisibility computed from all the pixels in the image. Given two frames $A$ and $B$ and the camera motion estimate between them \smalleq{$(\tensor[_B]{\ve{R}}{}{}^{{A}}, \ve{t}_B^A)\in\mathbb{SE}(3)$}, we transfer pixels from $A$ to $B$ using \eqref{eq:3d_registration} and compute the following pixel subsets:

\small

\begin{align}
 S_{\bar{H}} &= \left\lbrace\ve{p}\in\Omega \mid \mathcal{W}_A(\ve{p}) \neq \emptyset \right\rbrace, \\[8pt]
 S_{V} &= S_{\bar{H}}\cap\left\lbrace \ve{p}\in\Omega \mid (\ve{p}_B\in\Omega) \right. \notag \\
 &\phantom{= S_{\bar{H}}\cap\left\lbrace\right.}\left.\land  (\left|\mathcal{W}_B(\ve{p}_B)- w_B\right|<3\sigma_{\mathcal{W}}) \right\rbrace.
\end{align}

\normalsize

\smalleq{$S_{\bar{H}}$} is the subset of pixels in frame $A$ with valid inverse depth values (no-hole pixels) and 
\smalleq{$S_{{V}}$} the set of visible pixels when transferred to frame $B$. A pixel is tagged visible if it is within 
the image bounds, and it is not occluded in frame $B$. A pixel is considered occluded if its mapped inverse depth value 
is above an uncertainty allowed by the sensor noise model. Then the covisibility ratio is computed:

\small

\begin{align}
 vr_{AB} = \frac{\left|S_{V}\right|}{\left|S_{\bar{H}}\right|}
\end{align}

\normalsize

\noindent This procedure is repeated switching the role of $A$ and $B$, and then we select the minimum ratio as the final covisibility ratio.

%
%

\section{RGB-ID SLAM system}\label{sec:slam_system}

\begin{figure}
\centering
\includegraphics[width=0.9\linewidth]{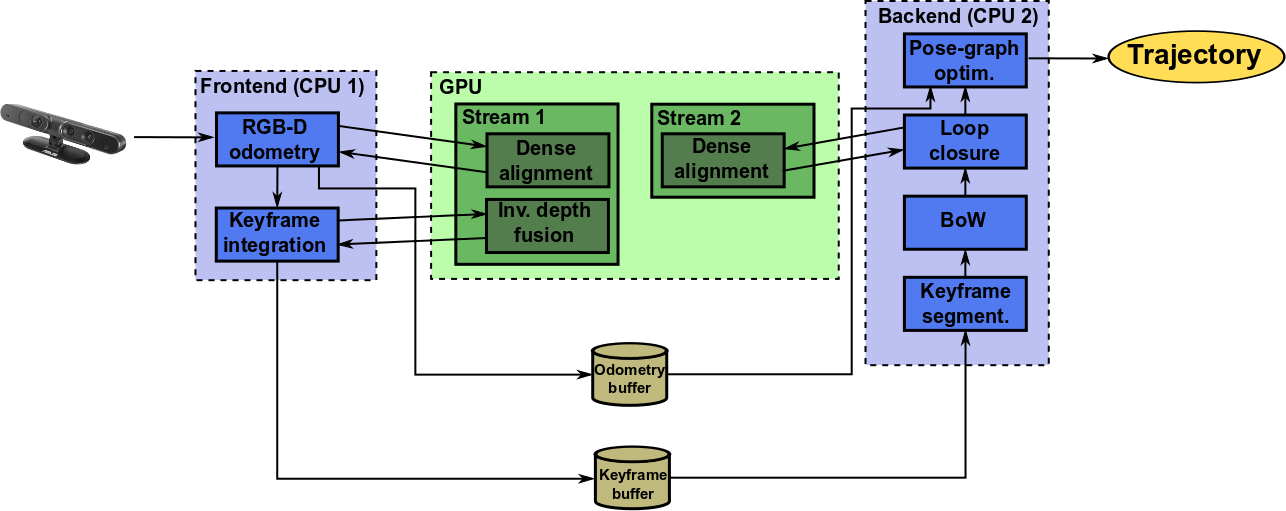}
\caption{Scheme of our complete RGB-ID SLAM system}
\label{fig:sysOverview}
\end{figure}

Our RGB-ID SLAM system is implemented in two CPU threads running concurrently. One thread executes a front-end for every incoming frame from the RGB-D sensor, performing camera tracking and fusion of inverse depth measurements in a single keyframe. For both of these tasks the CPU thread calls functions which execute in GPU where pixel-wise operations can be easily parallelised.  The second thread executes a back-end in charge of managing the keyframes passed by the front-end through a buffer. First it performs a segmentation of the keyframe in juts using the map of normals. Secondly for every keyframe it obtains different BoW histograms and stores the keyframe in a database. Finally it attempts to close loops comparing the last processed keyframe against the keyframes in the database. All the tasks of the back-end thread are performed in CPU except for a dense alignment in GPU between keyframes when a successful loop is detected. In order to allow for simultaneous access to GPU by both threads without 
blocking, each thread calls its GPU functions on its own CUDA stream which is executed asynchronously.

\subsection{Camera tracking}

For every upcoming frame \smalleq{$\left\lbrace\mathcal{I}_{k},\
\mathcal{W}_{k}\right\rbrace$} we have to compute the rigid body transformation \smalleq{$\tensor[_{rf}]{\ve{\hat{T}}}{}{}^{{k}}$} which best aligns it with a given reference frame \smalleq{$\left\lbrace\mathcal{I}_{rf},\ \mathcal{W}_{rf}\right\rbrace$}. This motion is computed by the dense RGB-D alignment method described in Section \ref{sec:dense_alignment}.
We provide an initial guess of the motion estimate using a constant velocity model, \ie, \smalleq{$\tensor[_{rf}]{\ve{\hat{T}}}{}_{(0)}^{{k}} = \tensor[_{rf}]{\ve{\hat{T}}}{}{}^{{k-1}}\tensor[_{k-2}]{\ve{\hat{T}}}{}{}^{{k-1}}$}.

Once we compute the motion estimate between the reference and current frame \smalleq{$\tensor[_{rf}]{\ve{\hat{T}}}{}^{{k}}$} and its covariance \smalleq{$\bm{\Sigma}_{rf}^{rf\rightarrow{k}}$}, we compute the sequential odometry constraint as it will be required in the latter pose-graph optimisation step.

\small


\begin{align}
 \tensor[_{k-1}]{\ve{T}}{}^{{k}} &= \left(\tensor[_{rf}]{\ve{T}}{}^{{k-1}}\right)^{-1}\tensor[_{rf}]{\ve{T}}{}^{{k}} \\
 \notag \\
 \bm{\Sigma}_{k-1}^{k-1\rightarrow{k}} = &
 \tfrac{\partial\tensor[_{k-1}]{\ve{T}}{}^{{k}}}{\partial\tensor[_{rf}]{\ve{T}}{}^{{k-1}}}\bm{\Sigma}_{rf}^{rf\rightarrow{k-1}} 
 \left(\tfrac{\partial\tensor[_{k-1}]{\ve{T}}{}^{{k}}}{\partial\tensor[_{rf}]{\ve{T}}{}^{{k-1}}}\right)^T + \notag \\
 &+ \tfrac{\partial\tensor[_{k-1}]{\ve{T}}{}^{{k}}}{\partial\tensor[_{rf}]{\ve{T}}{}^{{k}}}\bm{\Sigma}_{rf}^{rf\rightarrow{k}} 
 \left(\tfrac{\partial\tensor[_{k-1}]{\ve{T}}{}^{{k}}}{\partial\tensor[_{rf}]{\ve{T}}{}^{{k}}}\right)^T
\end{align}

\normalsize

\noindent where the computation of the Jacobians of the transformations is detailed in the Appendix \ref{app:covprop}. 



\subsection{Keyframe fusion}

After the visual odometry, the inverse depth map of current frame  is fused in the last selected keyframe. The criteria for keyframe selection is the same covisibility criteria as used for the reference frames taken for odometry. We must note that generally keyframes do not have to coincide with the odometry reference frames, since we allow for using a different covisibility threshold. Normally it will be set to a lower value for keyframe switching, \ie, we will initialise new keyframes at a lower rate than reference frames, using a visibility threshold of $0.7$. Once a keyframe is initialised, the inverse depth maps of the incoming frames are fused with the keyframe's. We not only fuse the incoming frames but also the frames recorded prior to the keyframe selection in order to integrate inverse depth measurements in parts of the image no longer observed after keyframe initialisation. To do so, frames are stored temporally in a buffer and erased once integrated in the later keyframe. Every time we integrate 
an incoming frame, we 
integrate also the frame closest in 
time of the ones remaining in the buffer.

Fusion in keyframes is performed in GPU using the reverse warping approach detailed in section \ref{sec:warping2}, where the frame to be integrated is warped towards the keyframe. Note that since we are using a reverse warping  scheme parts of the scene with missing depth measurements in the frame from which the keyframe is initialised cannot be reconstructed from subsequent frames. We find this a minor disadvantage given first, that reverse warping in CUDA is computationally cheap and easy to implement and secondly, that the generation of many keyframes generally guarantees that zones missing in one keyframe can be observed in another one.

Once the frame has been warped we have to integrate every new inverse depth value for every pixel in the keyframe. To do this we first verify that the pixels on both the keyframe and the warped frame correspond to the same scene point by performing the same covisibility check as in Section \ref{sec:covisibility_ratio}. If a pixel passes this test we update its inverse depth value \smalleq{$\mathcal{W}_{kf}(\ve{p})$} as well as the corresponding weight \smalleq{$\mathcal{C}_{kf}(\ve{p})$} in the keyframe as follows:

\small 

\begin{align}
 \mathcal{W}_{kf}(\ve{p}) &\leftarrow \frac{\mathcal{W}_{kf}(\ve{p})\mathcal{C}_{kf}(\ve{p})+\mathcal{C}_{k}(\ve{p}_w)\mathcal{W}_{k}(\ve{p}_w)}{\mathcal{C}_{kf}(\ve{p}) + \mathcal{C}_{k}(\ve{p}_w)} \\
 \notag \\
 \mathcal{C}_{kf}(\ve{p}) &\leftarrow \mathcal{C}_{kf}(\ve{p}) + \mathcal{C}_{k}(\ve{p}_w)
\end{align}

\normalsize

The weight \smalleq{$\mathcal{C}_{k}(\ve{p}_w) =\left( \frac{(1-\mathcal{W}_B(\ve{p}_w)\ve{e}_{\ve{z}}^T\ve{\tilde{t}}_B^A)^2}{\ve{e}_{\ve{z}}^T\tensor[_{B}]{\ve{{\tilde{R}}}}{}^{{A}}\ve{p}}\right)^2$} stems from the propagation of the inverse depth variance through \eqref{eq:3d_registration}. When a new keyframe is initialised the values of the weight map \smalleq{$\mathcal{C}_{kf}$} is reset to $1$ and the last keyframe is stored in a buffer, awaiting to be processed by the back-end in the second CPU thread.

\subsection{Keypoint extraction and BoW histograms}

Loop detection primarily relies on the DBoW2 place recognition system based in Bags of Binary Words proposed by G\'alvez-L\'opez and Tard\'os \cite{Galvez2012}. Mur-Artal et al. \cite{MurArtal2014} used this approach with ORB descriptors obtaining a quite reliable and fast loop closing system. As the authors suggest, we compute a pyramid of $8$ levels and a relative scale of $1.2$ between levels from the intensity image, and then we extract FAST keypoints at each pyramid level.  In order to distribute the keypoints uniformly over the image we divide the image in cells and set the firing threshold for the FAST extractor adaptively to extract the desired number of points at each cell. Once the keypoints are extracted, they are lifted to 3D points using the inverse depth map of the keyframe and ORB descriptors are computed in the intensity image. For the classification of the ORB descriptors in a BoW we use the same dictionary as in \cite{MurArtal2014}. For each keyframe a BoW histogram is computed with the 
frequency of each word in the given image. Finally the processed keyframe is stored in a keyframe database from which keyframes are queried in the loop closure process.

\subsection{Superjuts and entropy of normals}

For every keyframe which is passed to the back-end we generate a map of the degree of disarrangement of the elements in 
the scene,based on the entropy of the normals. For a given keyframe, we generate a normal map from its inverse depth 
map. Then we perform an object-like segmentation based on the algorithm for RGB images developed by Felzenszwalb and 
Huttenlocher~\cite{Felzenszwalb2004}, and which is similar to the methods presented in \cite{Triebel2010} and 
\cite{Karpathy2013}. The main difference is that we perform the segmentation directly on the keyframe pixels rather than 
on a lifted point cloud.

A graph $G=(V,E)$ is constructed, where each vertex in $V$ represents a
$3$D point $\ve{X}_i$ lifted from a pixel location and the edges $E$ represent the neighbouring relations between points. Edges are inserted between points with adjacent pixel locations on the image. Having the normals
at every point, a weight $w_{ij}$ for the edge joining vertices $i$ and $j$ is computed:

\small

\begin{align}
\renewcommand\arraystretch{2}
 w_{ij} = 
\left\{
\begin{array}{l} 
(1-\ve{n}_i^T\ve{n}_j)^2 \ \ \ \ \    \operatorname{if} \ \ve{n}_j^T(\ve{X}_j-\ve{X}_i) > 0 \\
 1-\ve{n}_i^T\ve{n}_j  \ \ \ \ \  \ \ \ \ \  \operatorname{if} \ \ve{n}_j^T(\ve{X}_j-\ve{X}_i) \leq 0 
\end{array}
\right.
\end{align}

\normalsize

\noindent where the squared weight is applied to convex edges, reflecting the fact that convex regions usually contain
points belonging to the same object and concave regions are likely to arise in frontiers between objects. After
computing the weights, the segmentation algorithm is run and essentially groups points sharing edges with low
weights in the same superjut. A parameter $k$ must be tuned such that the higher $k$,  the
larger segments will be obtained. We set this value always to $0.6$.

After the segmentation we compute a histogram of normals $F(\ve{n})$. Each of the $M$ bins in the histogram corresponds to a Voronoi cell associated to one point out of $M$
points uniformly distributed over the sphere. Since there is no
analytical solution to the problem of evenly distributing $M$ points on the sphere for any $M$, we use a simple approximate
solution based on the golden section spiral~\cite{Saff1997, Boucher2006}, which results in a discretisation with bins
covering areas of similar size. Finally given the histogram of normals of a jut, we can compute its entropy: 

\small 

\begin{align}
ent(F) = {\sum_{i=1}^M}F(\ve{n}_i)\log(F(\ve{n}_i)).
\end{align}

\normalsize

\subsection{Loop closure}

After extracting keypoints and computing the BoW histogram the keyframe database is searched for potential keyframe candidates for loop closures. Given two keyframes, DBoW2 returns a similarity score based on the distance between their BoW histograms, which we normalise with the score between the currently querying keyframe and the previous keyframe. If this score is above a threshold for any of the histogram comparisons a potential loop candidate is stored. In order to ensure that potential loop closures are obtained between temporally distant keyframes we establish a minimum keyframe separation.

For every possible keyframe candidate we  match its $3$D points with the $3$D points of current keyframe using their ORB descriptors. Then we perform a geometric validation of the detected loop by computing a rigid motion estimate applying the method described in \cite{Horn88} for alignment of 3D point clouds in a $3$-point RANSAC scheme. If more than $10$ points agree with the computed motion between keyframes, and the convex hull spanned by these points takes up more than $5\%$ of the image, the loop is tagged as correct. In order to compute the loop constraint, containing both a motion estimate and its covariance, we align both keyframes using the approach described Sec. \ref{sec:dense_alignment} where we pass the RANSAC rigid motion estimate as initial guess.

\subsubsection{Pose-graph optimisation}

When a new loop is detected it is added to a cluster containing loop constraints which are temporally close, allowing a maximum separation of $10$ keyframes between any two of the keyframes in the cluster. Pose-graph optimisation is applied when one of the clusters is no longer accepting new loop constraints, updating the graph with the new constraints in the cluster, as well as the poses and odometry constraints computed by the front-end, since the last graph update. Then, given a graph with a set of odometry constraints, \smalleq{$\mathcal{S} = \lbrace(1,2),...,(N-1,N)\rbrace$}, and a set of loop constraints, \smalleq{$\mathcal{R} =\lbrace(i_1,j_1),...,(i_L,j_L)\rbrace$}, we want to find the trajectory \smalleq{$\ve{x} = \left\lbrace \tensor[_{W}]{\ve{T}}{}^{{1}}, \ \tensor[_{W}]{\ve{T}}{}^{{2}}, \ ..., \ \tensor[_{W}]{\ve{T}}{}^{{N}}\right\rbrace$} which minimises the following cost function:

\small

\begin{align}
\renewcommand\arraystretch{2}
\begin{array}{l}
\ve{x}^*=\operatorname*{arg\,min}\limits_{\ve{x}} \sum\limits_{( i,j)\in\mathcal{S}}
{\ve{e}}^T_{ij}(\ve{x}){\bm{\Omega}_{i{j}}}{\ve{e}_{ij}(\ve{x})} +  \sum\limits_{(
i,j)\in\mathcal{R}} \ve{e}^T_{ij}(\ve{x}){\bm{\Omega}_{i{j}}}{\ve{e}_{ij}(\ve{x})}, 
\end{array}
\end{align}

\normalsize

\noindent where $\bm{\Omega_{ij}}$ is the information matrix of the constraint and the error term \smalleq{${\ve{e}_{ij}(\ve{x})}$} represents the residual between a constraint \smalleq{$\tensor[_{i}]{\ve{\hat{T}}}{}^{{j}}$} and the relative motion between the trajectory poses \smalleq{$\tensor[_{W}]{\ve{T}}{}^{{i}}$} and \smalleq{$\tensor[_{W}]{\ve{T}}{}^{{j}}$} it is connecting, expressed with a minimal parametrisation:

\small

\begin{align}
 \ve{e}_{ij}(\ve{x}) = 
  \log_{\mathbb{SE}(3)}\left(\tensor[_{i}]{\ve{\hat{T}}}{}^{{j}}(\tensor[_{W}]{\ve{T}}{}^{{j}})^{T}\tensor[_{W}]{\ve{T}}{}^{{i}}\right)
\end{align}

\normalsize

From this error term, the Jacobians and information matrix required for each constraints can be computed as detailed in 
Appendix \ref{app:posegraphjacs}.


\normalsize

To speed-up the process we perform the optimisation in a multi-layered scheme. In the first layer, odometry constraints between consecutive frames, are substituted by constraints between consecutive keyframes, which are also computed by the tracking front-end, and thus we only optimise the poses of the keyframes. In the second level the keyframe poses are fixed, and the poses of the rest of the frames are optimised by enforcing the odometry constraints.

\section{Experiments}\label{sec:experiments}

We first compare our method quantitatively with the state-of-the-art in terms of trajectory estimation. Then we perform a qualitative evaluation of the 3D reconstructions both in the Freiburg and our own RGB-D sequences, and finally we measure the computational performance, both of the front-end, in charge of camera tracking, and the back-end, performing keyframe processing and loop closing.

      \begin{table*}
      \centering
      \caption{Absolute trajectory error (RMSE, median and max) in meters of our method with and without loop closure and comparison with state-of-the-art approaches. For a given error measure in a dataset, we show in bold the best approach in the state of the art and ours if it is the absolute best.}
      \scalebox{0.5}
      {
      \begin{tabular}{|c|c|c|c|c|c|c|c|c|c|c|c|c|c|c|c|c|c|c|}
      \hline
       Approach &  \multicolumn{3}{c|}{fr1/desk} & \multicolumn{3}{c|}{fr1/desk2} & \multicolumn{3}{c|}{fr1/room} & \multicolumn{3}{c|}{fr2/desk} & \multicolumn{3}{c|}{fr3/office} & \multicolumn{3}{c|}{fr3/nst}\\
      \cline{2-19}
      & RMSE & median & max & RMSE & median & max & RMSE & median & max & RMSE & median & max & RMSE & median & max & RMSE & median & max\\
      \hline
      Ours (w/o loop closure)		 & {0.034} & 0.030 & 0.087 		& 0.054 & 0.037 & 0.137 	  & {0.087} & 0.072 & 0.203 			 & {0.037} & 0.026 & 0.084 &	 0.057 & 0.042 & 0.118							   & 0.041 & 0.030 & 0.106	\\
      Ours (w/ loop closure)$*$ & \textbf{0.018} & \textbf{0.014} & \textbf{0.052} 		& {0.036} & 
\textbf{0.029} & \textbf{0.104} 	  & {\textbf{0.040}} & \textbf{0.035} & \textbf{0.140} 			 		
	   & {0.017} & \textbf{0.015} & \textbf{0.036} 	& 0.025 & \textbf{0.024} & \textbf{0.046}	& \textbf{0.015} 
& 0.012 & 0.051\\
      \hline
      \cite{GutierrezGomez2016} & 0.033 & 	{0.026} & 	0.086 & 	{0.066} & 	{0.044} & 	{0.180} & 	0.097 & 	0.086 & 	0.195 & 	{0.075} & 	0.077 & 	 f{0.104} & 0.082 & 	0.036 & 	0.143 & - & - & - \\
      Unified VP \cite{Meilland2013b} 			& - & \textbf{0.018} & \textbf{0.066}		 & - & - & -		
  & -     & 0.144 & 0.339		 & - & {0.093} & {0.116} & - & - & - & - & - & -\\
      {ICP+RGB-D \cite{Whelan2013ICRA}} 	& - & 0.069 & 0.234		 & - & - & -		  & -     & 0.158 & 0.421		 & - & 0.119 & 0.362 & - & - & -  & - & - & -\\
      {6D RGB-D odometry \cite{Dong2014}}		& - &     - & - 		  & - & - & -		  & 0.095 & \textbf{0.067} & 0.254		 & 0.197 & 0.174 & 0.416 & - & - & - & - & - & -\\
      {SDF tracking \cite{Bylow2013}  }       & 0.035 & - & - 		& 0.062 & - & - 	 & 0.078 & - & - 			& - & - & - & 0.040 & - & - & - & - & -\\
      DIFODO \cite{Jaimez2015} & {{0.047}} & - & - 		& {0.094} & - & - 	  & {{0.109}} & - & - 			 & {0.342} & - & - & - & - & - & - & - & -\\
      SLAC \cite{Zhou2014} & 0.026 & - & -	& \textbf{0.035} & - & - 	& 0.059 & - & -		& - & - & -	
& 0.022 & - & - & - & - & - \\
      ElasticFusion (w/o loop closure) \cite{Whelan2016} & {{0.022}} & - & - 		& - & - & - 	  & - & 
- & - 			 & - & - & - & 0.025 & - & - & 0.027 & - & -\\
      {RGB-D SLAM \cite{Endres2014}}$*$		&0.023 & - & - 			& {0.043} & - & - 	  & 0.084 & - & 
- 			 & 0.057 & - & - & 0.032 & - & - & 0.017 & - & -\\
      MRSMap \cite{Stuckler2012}$*$ 				& 0.043 & - & - 		& 0.049 & - & - 	  & 0.069 & - & - 			 & 0.052 & - & - & 0.042 & - & - & 2.018 & - & -\\ 
      {DVO-SLAM \cite{Kerl2013IROS}}$*$     & {0.021} & - & - 		& 0.046 & - & - 	  & \textbf{0.053} & - & - 			 & \textbf{0.017} & - & - & 0.035 & - & -& 0.018 & - & -\\
      Kintinious \cite{Whelan2014}$*$  & 0.037 & 0.031 & 0.078 		& 0.071 & - & - 	  & 0.075 & 0.068 & \textbf{0.231} 			 & 0.034 & \textbf{0.028} & \textbf{0.079} & {0.030} & - & - & {0.031} & - & -\\     
      ElasticFusion  (w/ loop closure) \cite{Whelan2016} $*$ & {\textbf{0.020}} & - & - 		& 0.048 & - & - 
	  & 0.068 & - & - 			 & 0.071 & - & - & \textbf{0.017} & - & - & \textbf{0.016} & - & -\\
      \hline
      \multicolumn{19}{l}{$*$ with loop closure and pose-graph optimisation (deformation graph for ElasticFusion)}\\
      \end{tabular}
      }
      \label{tab:ATE2}
      \end{table*}

\subsection{Trajectory Estimation}

We first compare our method to state-of-the-art visual odometry and SLAM approaches with RGB-D systems. For the comparison we use the TUM benchmarking dataset \cite{RGBDbenchmarkTUM}.  The evaluation has been carried out taking the error metric of the Absolute Trajectory Error (ATE) in meters (see Table \ref{tab:ATE2}).  However it is worth noticing that there exists a large improvement in the $fr2/desk$ dataset. Apparently the reason is a correction we applied to the depth maps of the Freiburg 2 sequences by a constant scaling factor. This correction was in principle reported as done by the authors of the dataset. However it seems from the observation made by Mur-Artal et al. in \cite{MurArtal2014} of a scale bias in the RGB-D SLAM system of 
Endres et al. \cite{Endres2014}, that it has not been applied in the Freiburg 2 datasets available for download.

The table shows that our method outperforms most of the odometry and SLAM methods in the literature and it is close to 
the ElasticFusion of Whelan et al. \cite{Whelan2016}, which shows the best performance in the state-of-the-art in the 
datasets $fr3/office$ and $fr3/nst$. Among the methods performing only odometry, ElasticFusion in tracking-only mode 
outperforms them and even some of the methods performing loop closure. However in other datasets our method shows 
overall the best performance both in tracking-only mode and with loop closures. It is worth noticing that challenging 
datasets like $fr1/desk2$ and $fr1/room$, showing a fast camera motion, our method provides a great precision.

\subsection{3D reconstruction}
\begin{figure}
\begin{tabular}{cc}
 \includegraphics[width=0.45\linewidth]{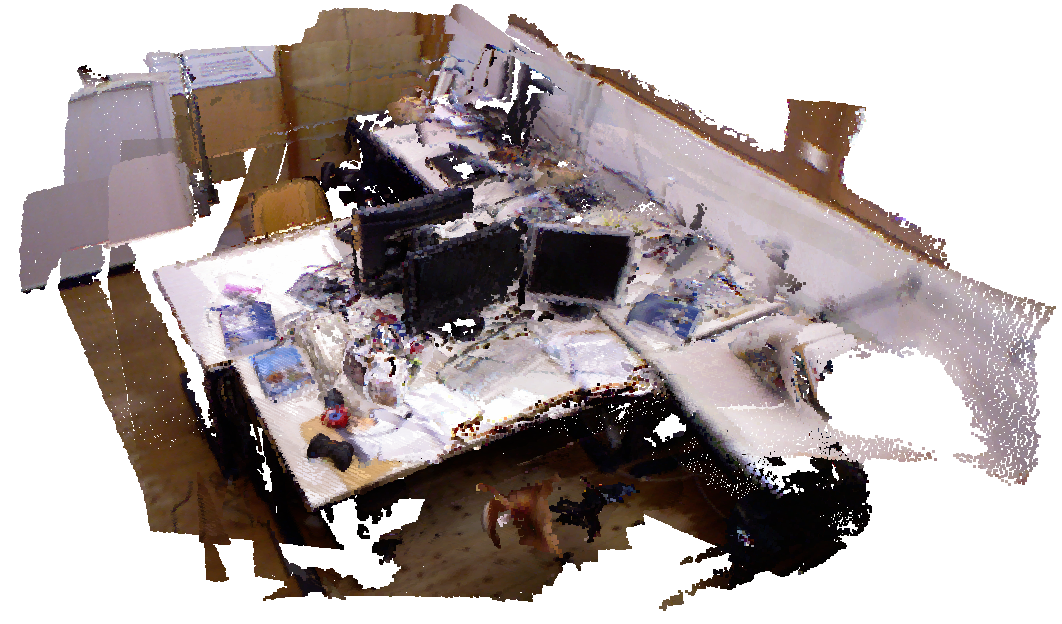}&
\includegraphics[width=0.45\linewidth]{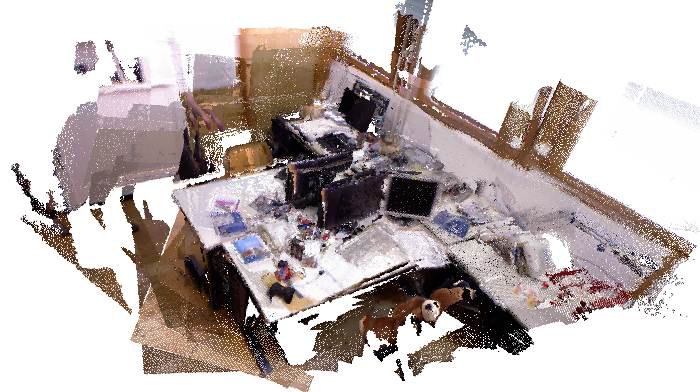}\\
fr1/desk & fr1/desk2 \\
\includegraphics[width=0.45\linewidth]{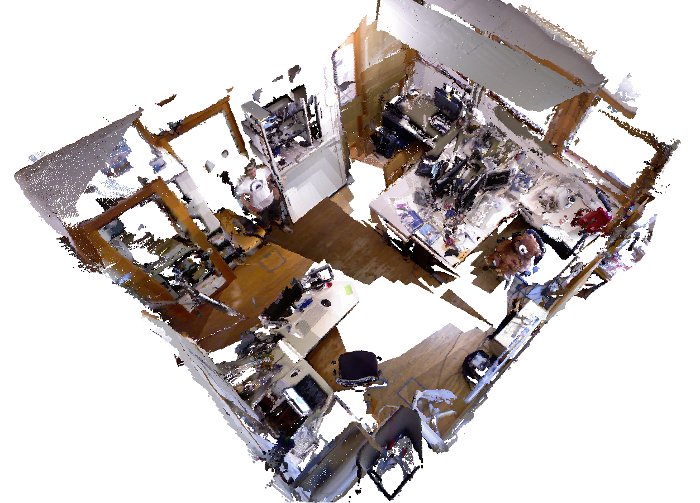}&
\includegraphics[width=0.45\linewidth]{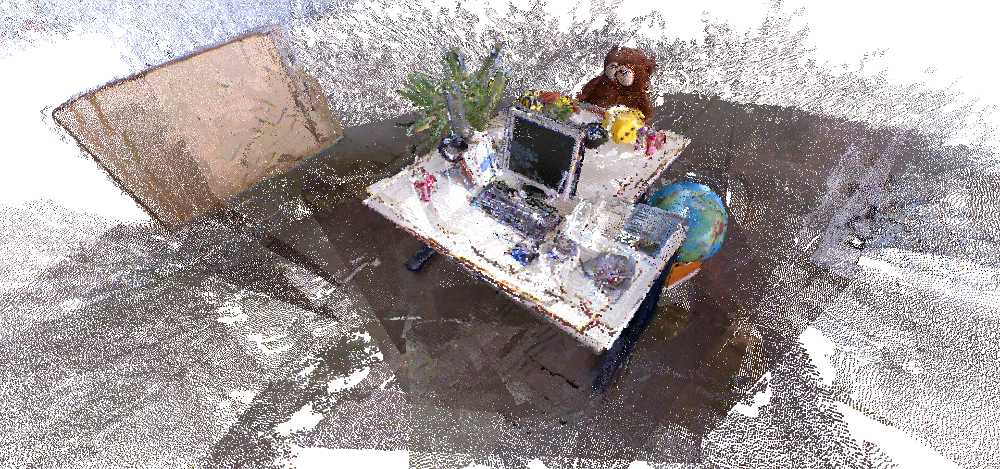}\\
fr1/room & fr2/desk \\
\includegraphics[width=0.45\linewidth]{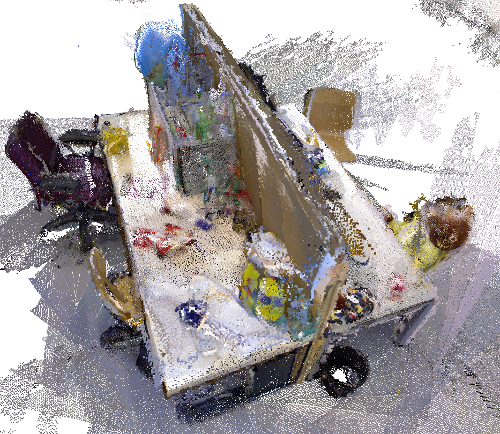}&
\includegraphics[width=0.45\linewidth]{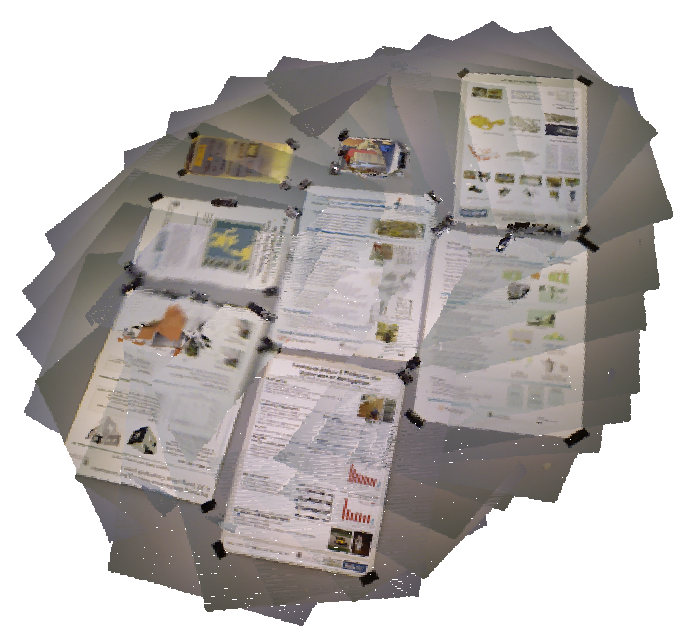}\\
 fr3/office & fr3/nst
\end{tabular}
\caption{Final 3D models from the TUM datasets on which our method has been evaluated. Despite the discontinuities in color, we can appreciate a good level of detail in the final reconstruction. }
\label{fig:3Dreconstructions_TUM}
\end{figure}

In this section we perform a qualitative evaluation of the 3D models reconstructed of our SLAM system. To build the 3D model of the scene we concatenate the point clouds of different keyframes in real-time. To avoid unnecessary redundant points we only concatenate partial point clouds which correspond to novel parts of the scene between consecutive keyframes. Redundant points due too loop closure are eliminated at the end of the execution of our method by performing a voxel grid filtering with a voxel size of $1$ cm. In Fig. \ref{fig:3Dreconstructions_TUM} we show the 3D models of the evaluated TUM datasets, showing a great detail of the reconstructions.

\subsection{Computational performance}

The computational performance of our system has been evaluated in a laptop PC with an Intel Core i7-4710HQ CPU at 2.50GHz, 16GB of RAM and a nVidia GeForce GTX 850M GPU with 4GB of memory. As shown in Table \ref{tab:execution_frontend}, the execution time of the front-end varies depending on whether the system operates in tracking only-mode, \ie, with the back-end disabled or with its complete functionalities. This is due to the division of the GPU resources upon successful loop closure between the visual odometry and integration modules from the front-end, and the keyframe alignment module between loop keyframes from the back-end.

\begin{figure}
\begin{tabular}{cc}
 \includegraphics[width=0.45\linewidth]{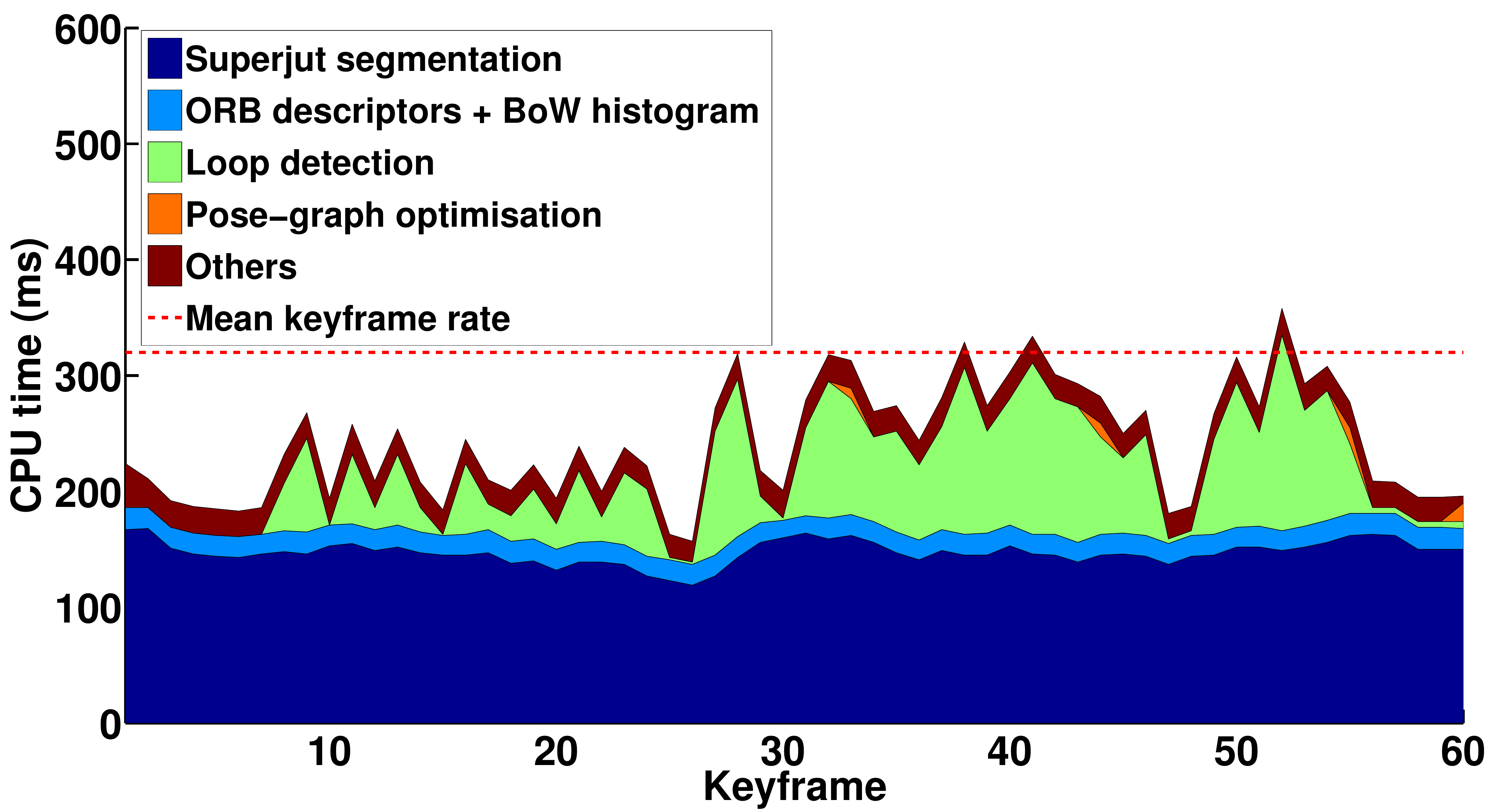}&
\includegraphics[width=0.45\linewidth]{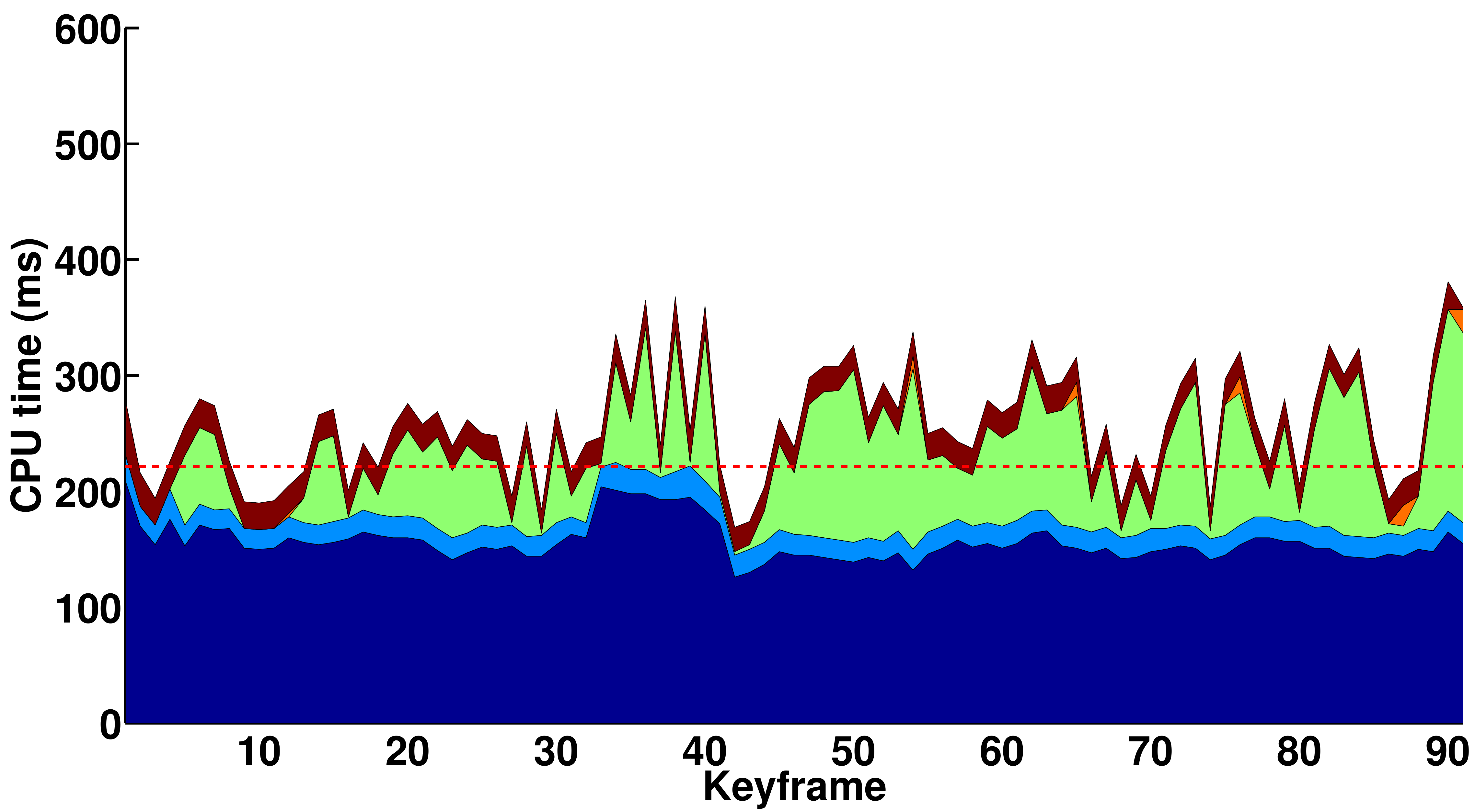}\\
fr1/desk & fr1/desk2 \\
\includegraphics[width=0.45\linewidth]{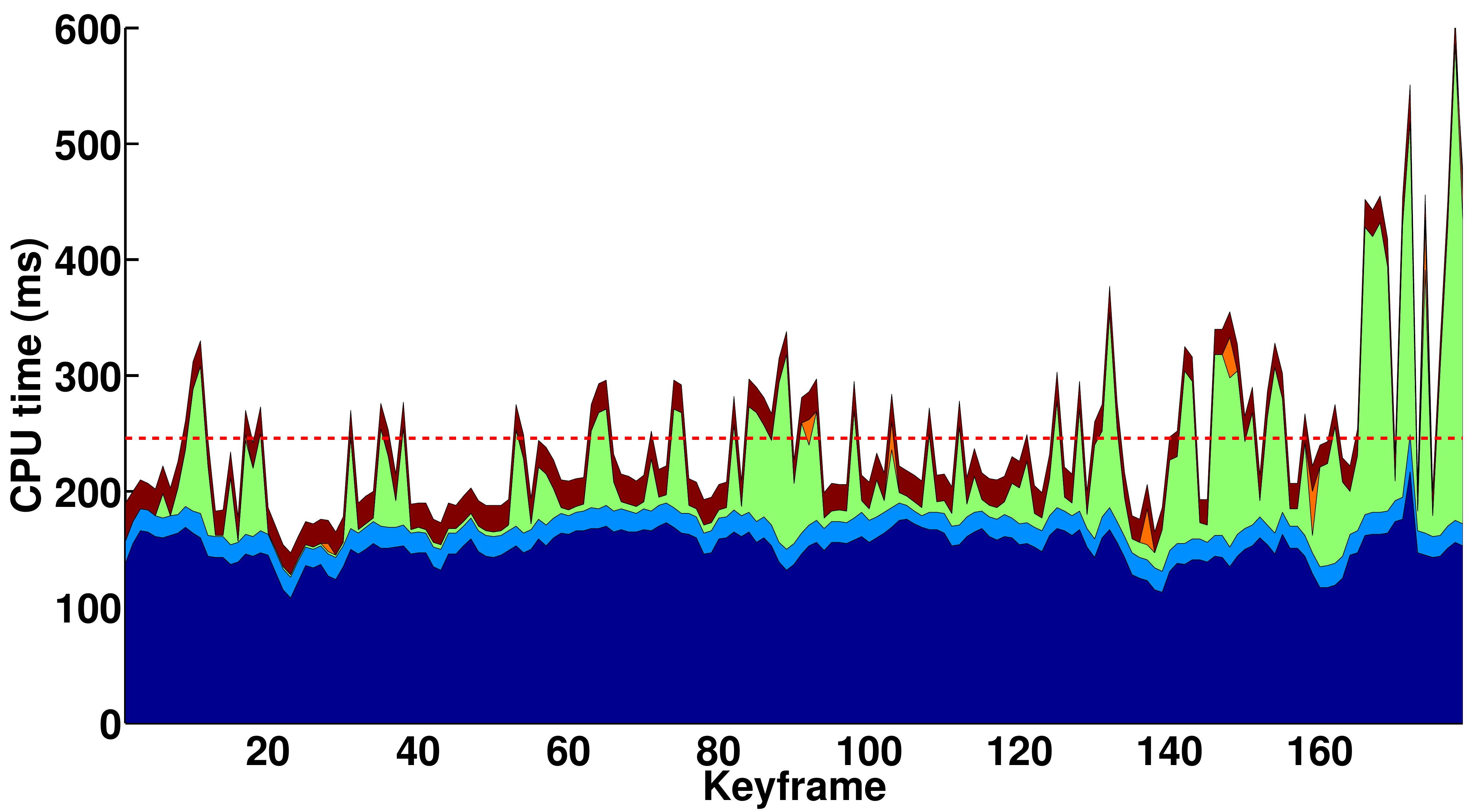}&
\includegraphics[width=0.45\linewidth]{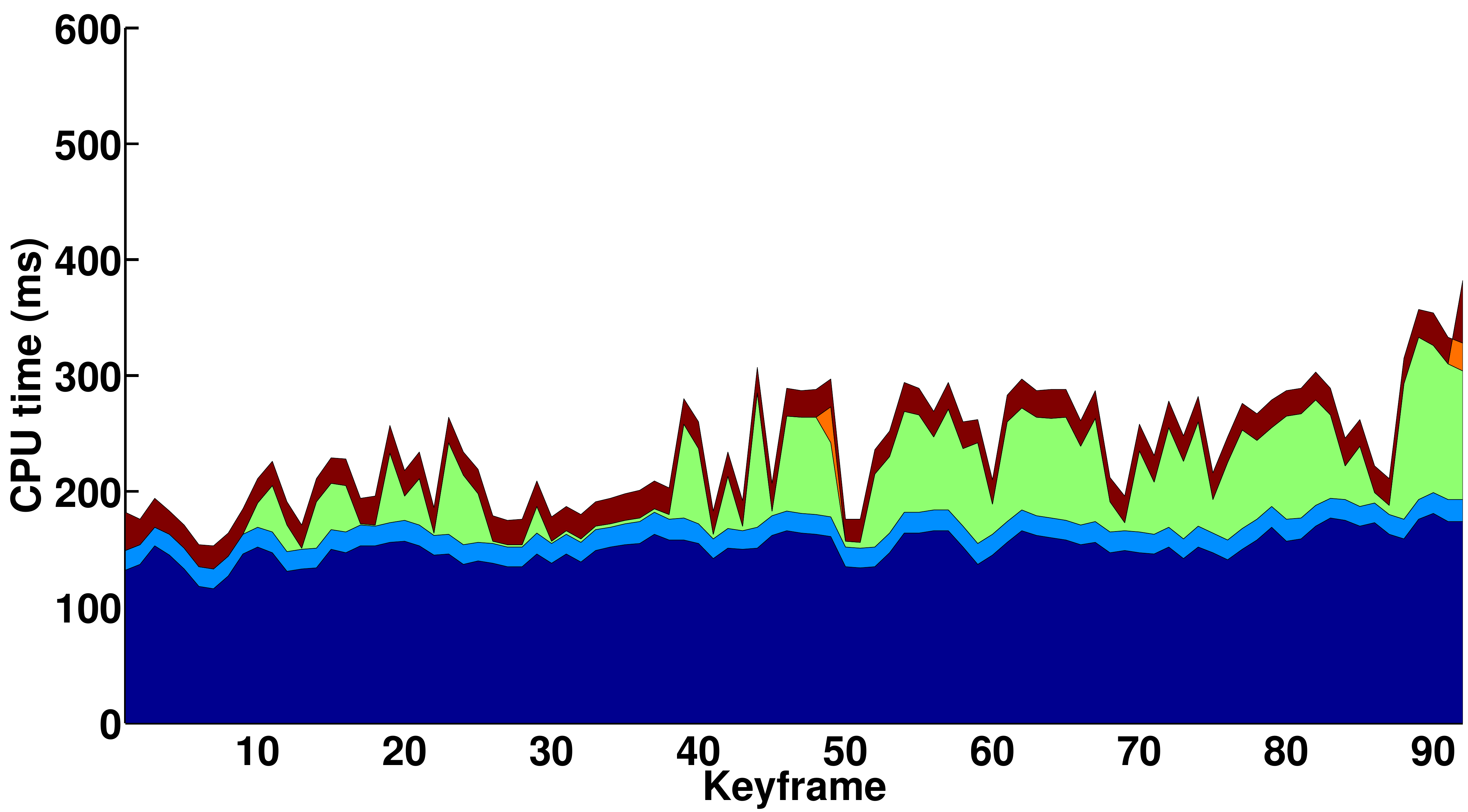}\\
fr1/room & fr2/desk \\
\includegraphics[width=0.45\linewidth]{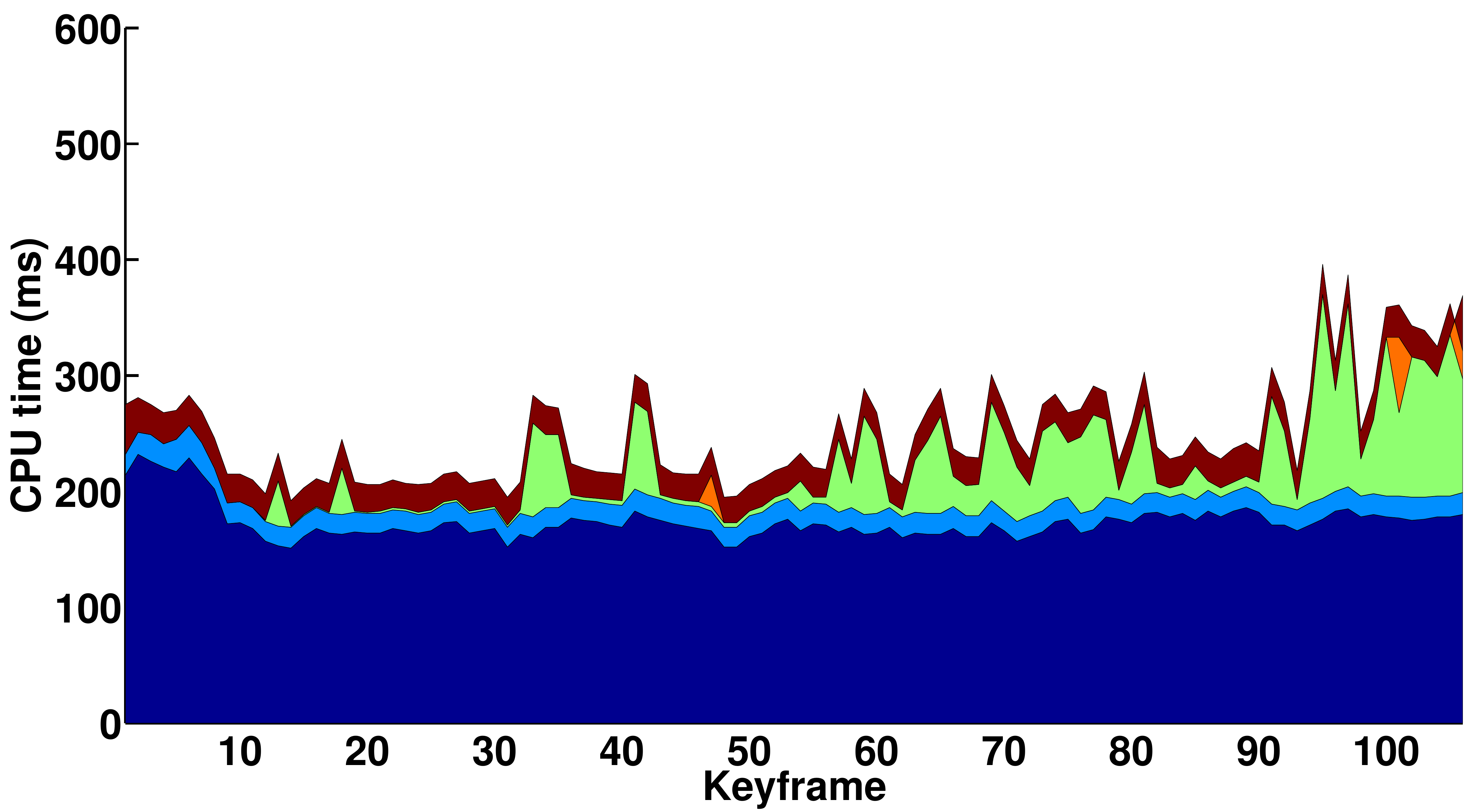}&
\includegraphics[width=0.45\linewidth]{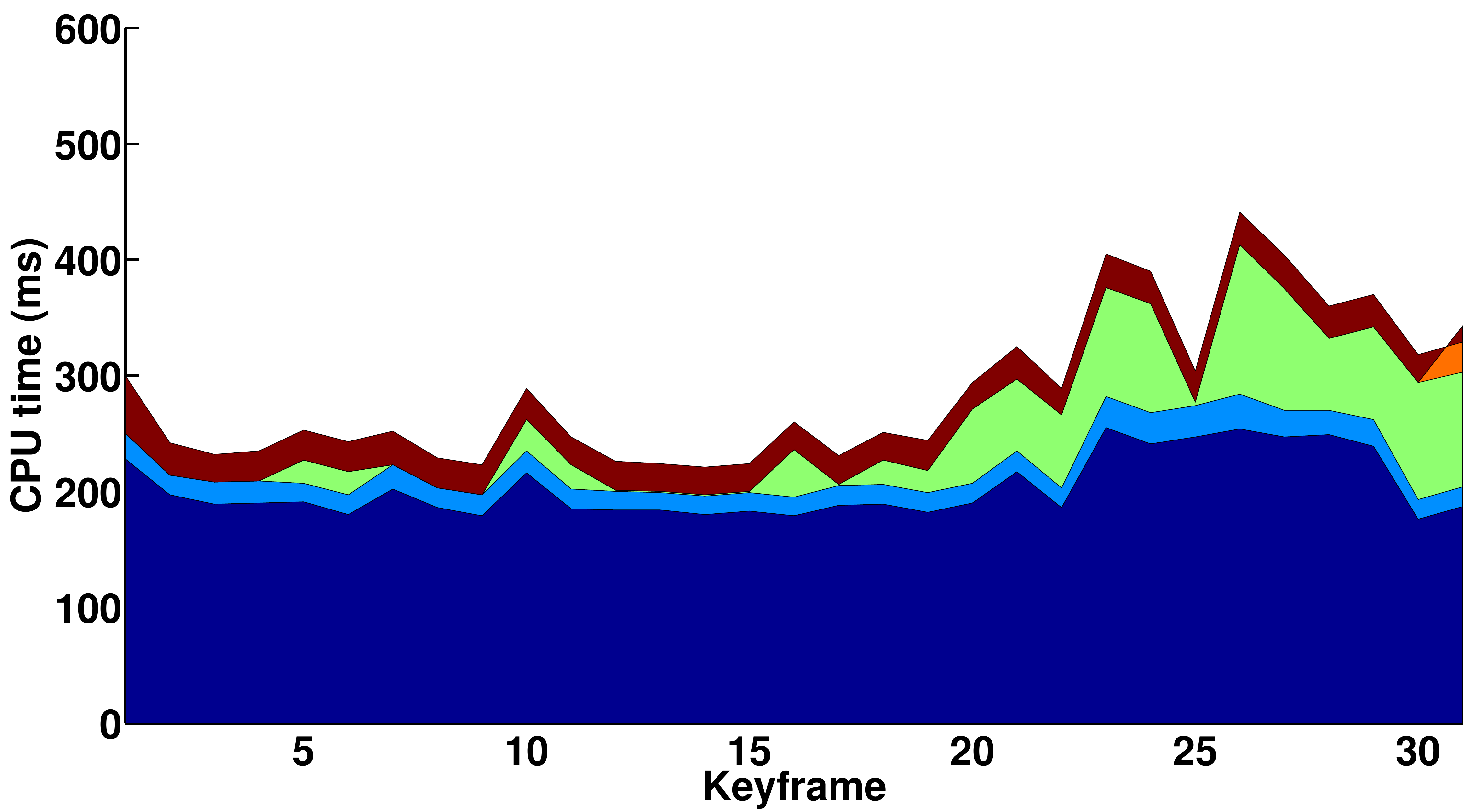}\\
 fr3/office & fr3/nst
\end{tabular}
\caption{Computational cost of the keyframe processing in the back-end divided by processes (area graph), and average acquisition time of a keyframe (dashed red). In the datasets where the average acquisition time is not shown, it means that it is out of the plot bounds (above 600 ms). This usually occurs in datasets when the camera executes a loop around some scene, which normally involves that keyframes are switched with low frequency.}
\label{fig:backend_cost}
\end{figure}

In Fig. \ref{fig:backend_cost} we show graphically the computational cost of the back-end pipeline, which has been dissected in the different processes involved. As we can observe there is a nearly constant cost for keyframe segmentation and descriptor extraction, while the cost for loop closure detection and pose-graph optimisation shows a great variability. The reason is that most of the cost of loop closure detection is governed by the step of geometric verification and loop constraint computation, which is run only on a reduced list of loop candidates delivered by the efficient appearance-based BoW algorithm for loop detection. Red dashed line represents the time it takes the front-end to deliver a new keyframe to the back-end, assuming that it is able to process al the frames in the sequence at the frame rate of $30$ Hz. We can observe that in datasets where the camera is moved slowly in a loop around the same scene, the back-end processing time is widely below the average keyframe rate. However in 
datasets like $fr1/desk2$ or $fr1/room$, where the camera moves quickly and/or the area being mapped changes a lot, the back-end struggles to operate at keyframe rate. These observations are embodied numerically in Table \ref{tab:execution_backend}.

\begin{table*}
      \centering
      \caption{Computational cost of the front-end pipeline in milliseconds}
      {
      \begin{tabular}{c|cccccc}      
        &  {fr1/desk} & {fr1/desk2} & {fr1/room} & {fr2/desk} & {fr3/office} & {fr3/nst}\\
      \hline
      w/o loop closure    &    39    &           39      &        42    &       38     &        38    &         36 \\
      w/ loop closure    &    46       &        55       &       76    &       49    &         53     &        41 \\
      \hline
      \end{tabular}
      }
\label{tab:execution_frontend}
\end{table*}

\begin{table*}
      \centering
      \caption{Mean computational cost of the processes involved in the back-end pipeline in milliseconds. To achieve online performance the mean total cost should be lower than the mean keyframe rate.}
      \
      {
      \begin{tabular}{c|cccccc|}      
        &  {fr1/desk} & {fr1/desk2} & {fr1/room} & {fr2/desk} & {fr3/office} & {fr3/nst}\\
      \hline
      Segmentation		 & 147	 & 156	 & 152	 & 151	 & 174	 & 203 \\
      ORB desc. + BoW hist.	 & 18	 & 19	 & 18	 & 17	 & 18	 & 19 \\
      Loop detection		 & 54	 & 61	 & 52	 & 46	 & 33	 & 36 \\
      Pose-graph optim.		 & 1	 & 1	 & 1	 & 1	 & 1	 & 1 \\
      \hline
      Total keyframe proc.	 & 243	 & 260	 & 246	 & 237	 & 250	 & 286 \\
      \hline
      Mean keyframe rate	 & 320	 & 222	 & 246	 & 1017	 & 773	 & 1735 \\
      \hline
      \end{tabular}
      }
\label{tab:execution_backend}
\end{table*}

\subsection{Sensor Calibration and Depth Correction}

\begin{figure}
\begin{tabular}{cc}
 \includegraphics[width=0.45\linewidth]{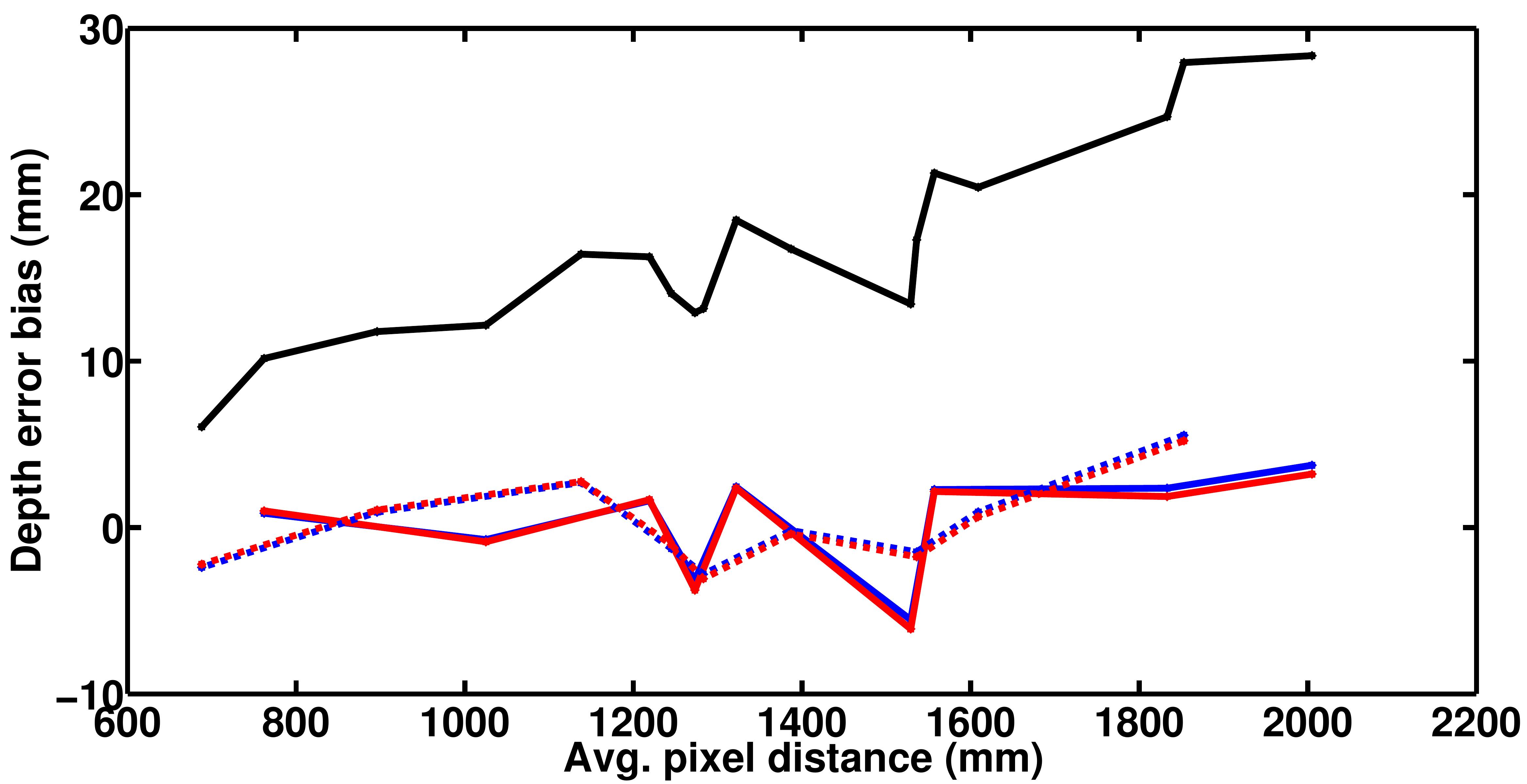}&
\includegraphics[width=0.45\linewidth]{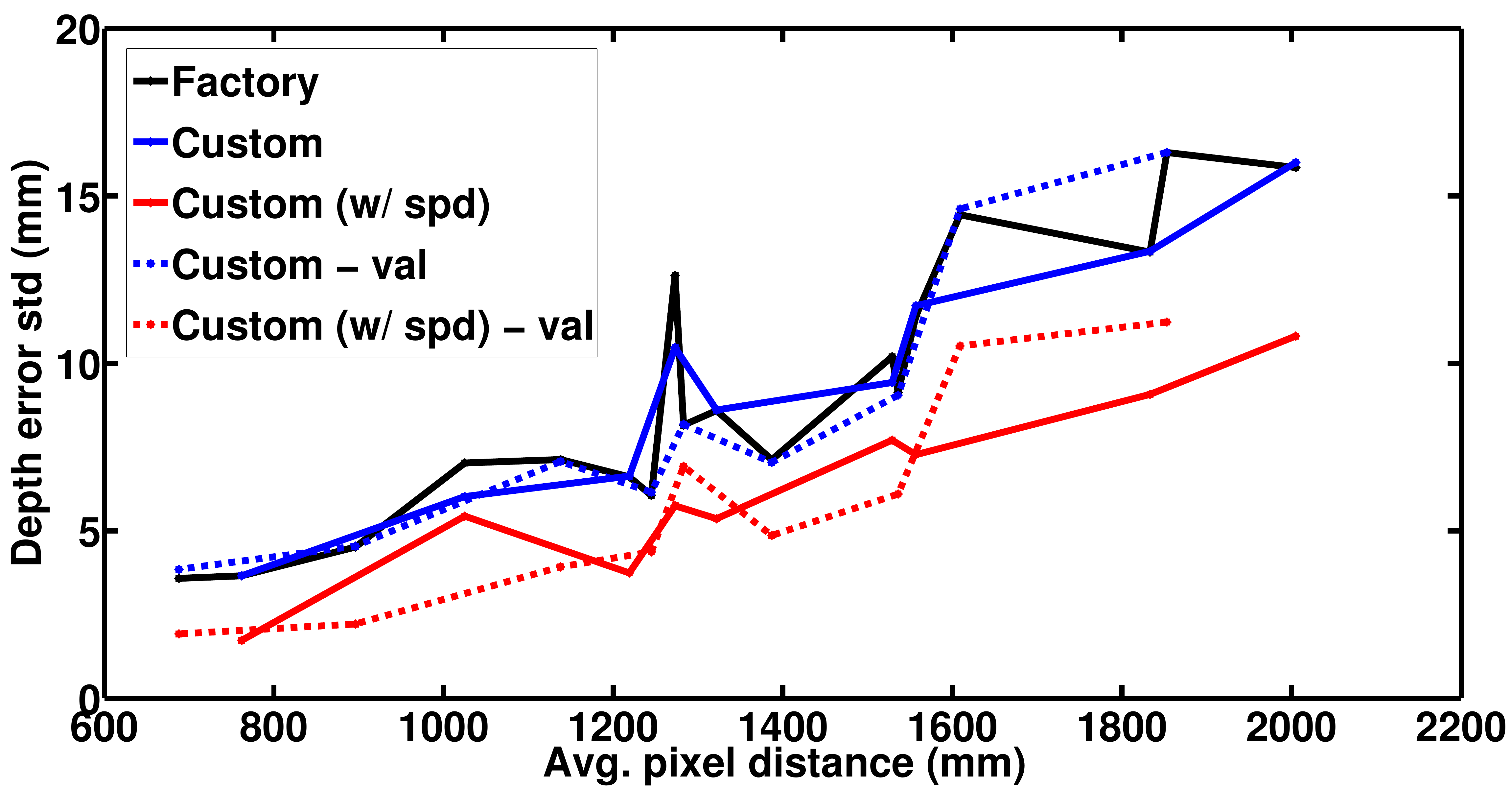}\\
 \includegraphics[width=0.45\linewidth]{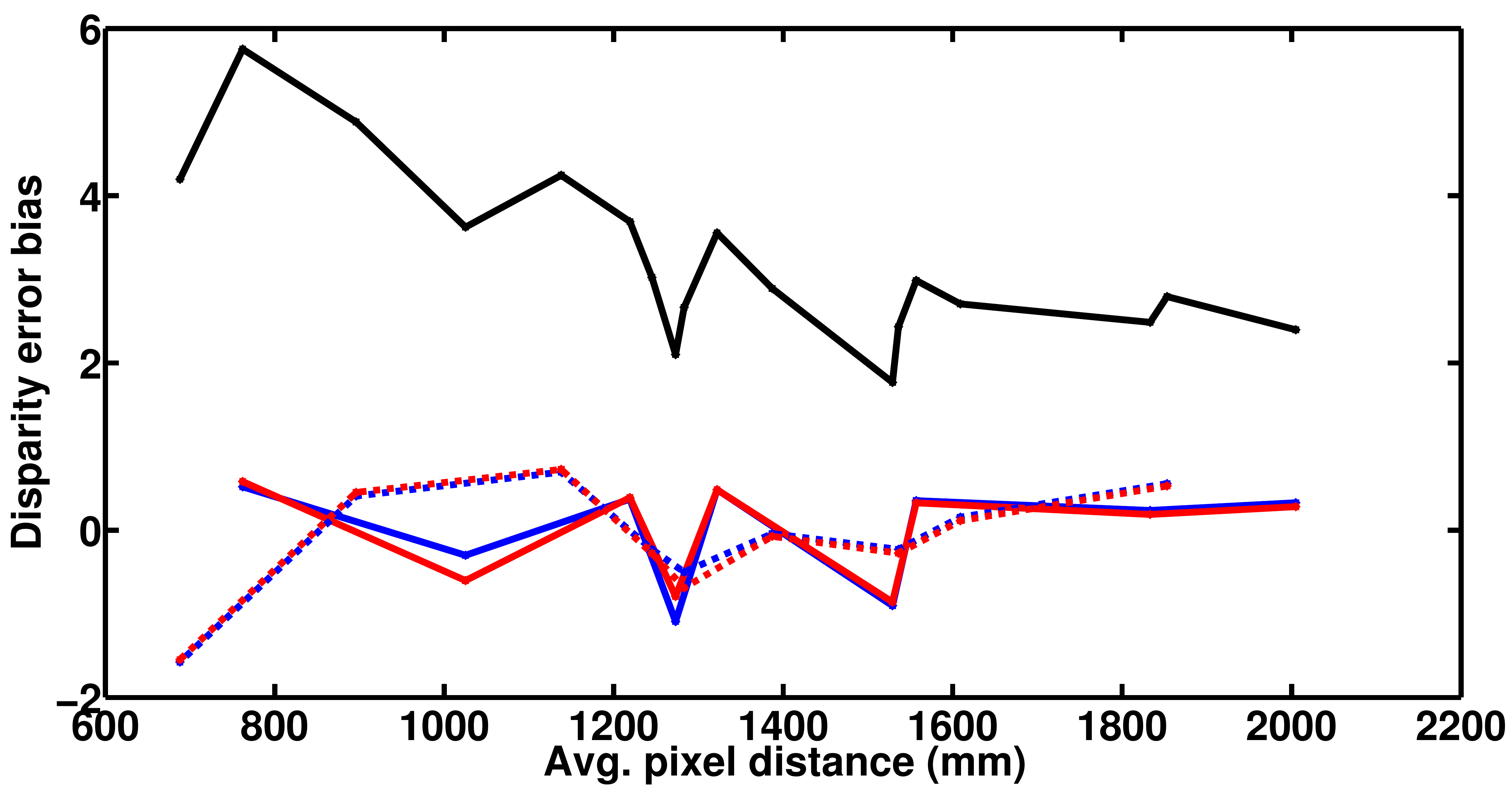}&
\includegraphics[width=0.45\linewidth]{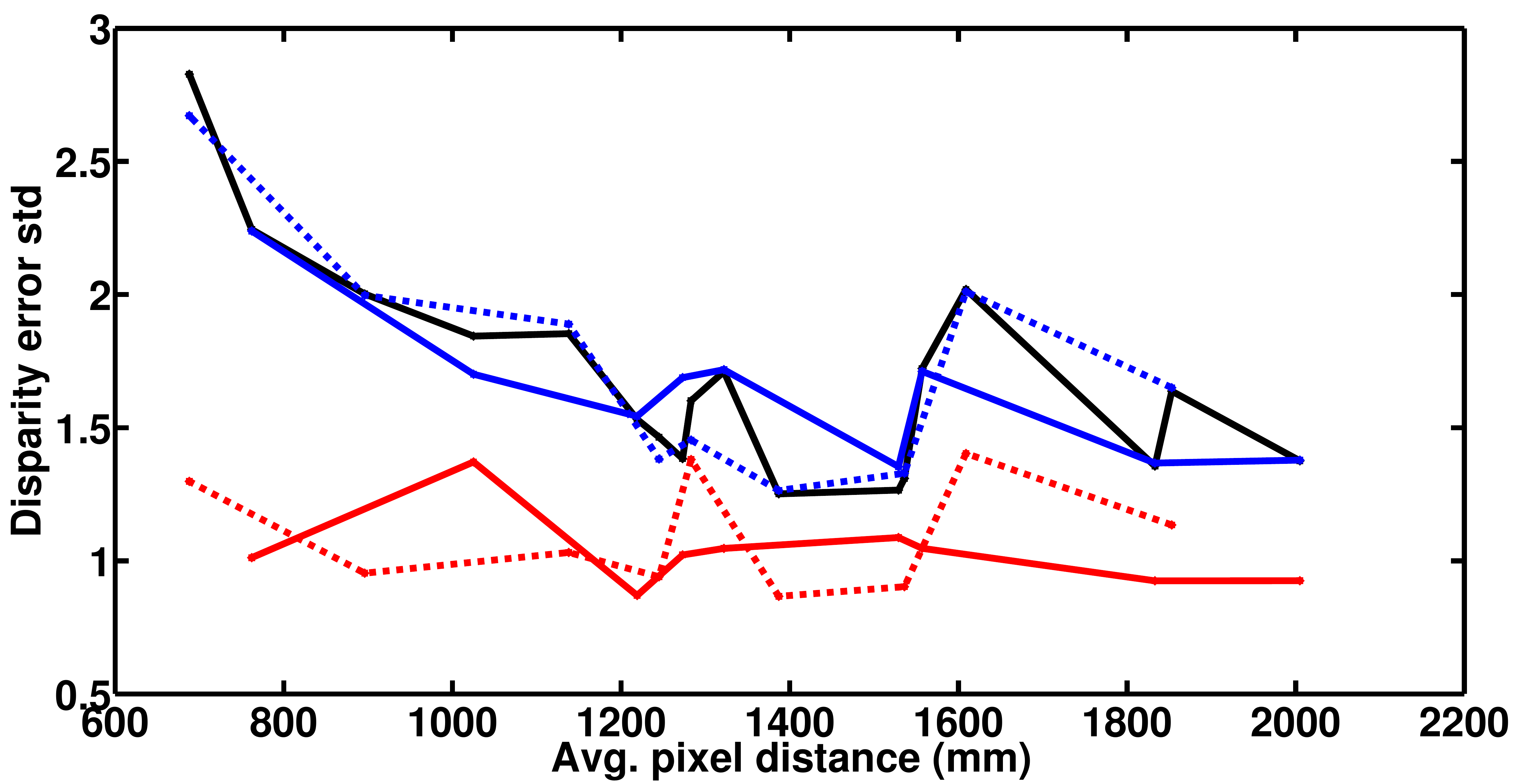}
\end{tabular}
\caption{Bias and standard deviation of the calibration error, varying with the average pixel distance of each image for different calibration methods. Error is shown in depth units (top) as well as in disparity units (bottom).}
\label{fig:calibration_error}
\end{figure}

In this section we evaluate the calibration model and the procedure proposed in Section \ref{sec:calibration} to compute a custom calibration for a specific sensor. The calibration is done in two steps. In the first step we take pairs of RGB and IR images compute the RGB and IR intrinsics as well as the rigid transformation between the RGB and depth reference frames, using the Bouguet Toolbox \cite{BouguetCalibration}. In the second step we calibrate the disparity-to-depth conversion of the depth sensor. To do this, we take $18$ shots of synchronised RGB and depth images  with the camera pointing a textured wall. From the RGB images we have generated a ground truth, by manually selecting and matching some dozens of points across images, and then performing a Bundle Adjustment with the Photomodeler software to obtain the camera poses as well as a 3D point cloud from which the best fitting plane is computed. Then, using \eqref{eq:plane_restriction} with the calibration parameters computed in the first step, 
we generate a dense inverse depth map which is used as ground truth in the optimisation function used for calibration of the disparity-to-depth conversion. To test the accuracy of our method we have selected the odd images for calibration and the even ones for validation. The comparison of our calibration, with and without depth spatial distortion correction, and the factory calibration is shown in Fig. \ref{fig:calibration_error}. We can observe in the figure, that there exist a bias in the raw depth map which is completely corrected by calibrating only the linear model. However, after this adjustment it still remains a zero-mean error with a spatial pattern, which is corrected by the calibrated model of the depth spatial distortion. Note that with a custom calibration the depth error is reduced in both the calibration and the validation images, which indicates that the disparity-to-depth conversion is well modelled by our calibration model. 

\begin{figure}
\includegraphics[width=0.95\linewidth]{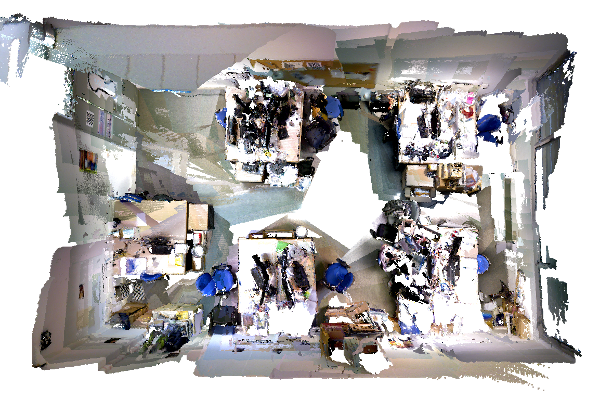}
\caption{3D reconstruction obtained by our RGB-iD SLAM system in an office environment using a custom calibration for the RGB-D camera, and with unregistered depth images. }
\label{fig:labReconstruction}
\end{figure}

To evaluate the performance of our RGB-iD SLAM system with a custom calibration we have run our method on one sequence 
taken with synchronised RGB and depth streams but with unregistered depth maps. The obtained reconstruction is shown in 
Fig. \ref{fig:labReconstruction}

\section{Conclusions}\label{sec:conclussions}

In this paper we have presented a complete RGB-ID SLAM system which combines an accurate dense and direct RGB-D odometry 
system and state-of-the-art loop closing procedure to achieve a great accuracy in the estimation of the trajectory. The 
algorithm also performs the integration of the tracked frames into keyframes with smooth inverse depth maps. The fusion 
of depth measurements in keyframes also allows to achieve a greater accuracy in the computation of relative camera 
transformation at loop closure. The concatenation of the novel parts between consecutive keyframes allow to obtain 
precise 3D models of the environment. In the comparison with some of the best state of-the-art methods for RGB-D mapping 
our method shows to have results with greater or similar accuracy in most of the tested datasets. Our method is able to 
work in real time with a mid-range laptop GPU. Our system is also to work with custom calibrated RGB-D sensors without 
need of hardware registering the depth image to the RGB frame. With this 
in mind we have also proposed a pipeline for accurate calibration of RGB-D cameras. The code for the RGBiD-SLAM system 
has been made available publicly \footnote{https://github.com/dangut/RGBiD-SLAM}.

\appendix 

\section{Covariance propagation in composition of spatial transformation}\label{app:covprop}
Let us define the following composition operation of spatial transformations \smalleq{$\tensor[_{A}]{\ve{{T}}}{}^{{B}} = 
\left(\tensor[_{W}]{\ve{{T}}}{}^{{A}}\right)^{-1}\tensor[_{W}]{\ve{{T}}}{}^{{B}}$}. Let us suppose that 
\smalleq{$\tensor[_{W}]{\ve{{T}}}{}^{{A}})$} and \smalleq{$\tensor[_{W}]{\ve{{T}}}{}^{{B}}$} have covariances 
\smalleq{${\bm{\Sigma}}_W^{WA}$} and \smalleq{${\bm{\Sigma}}_W^{WB}$} respectively expressed in the reference $W$, and 
we want to compute the covariance  \smalleq{${\bm{\Sigma}}_A^{AB}$}. To compute the Jacobians we must apply differential 
perturbations in the side in which the covariance is referenced. In this case all of them are left perturbations. Taking 
separately the translation and rotation we have:

\footnotesize

\begin{align}
\ve{t}_A^B+\bm{\delta}\ve{t}_A^{AB} =  \left(\exp\left({\bm{\delta}\bm{\theta}_{W}^{WA}}\right)\tensor[_{W}]{\ve{{{R}}}}{}^{{A}}\right)^{T}\left(\ve{t}_W^B+\bm{\delta}\ve{t}_W^{WB}-\ve{t}_W^A-\bm{\delta}\ve{t}_W^{WA}\right)\label{eq:translation_pert}\\
 \exp\left({\bm{\delta}\bm{\theta}_{A}^{AB}}\right)\tensor[_{A}]{\ve{{{R}}}}{}^{{B}} = \left(\exp\left({\bm{\delta}\bm{\theta}_{W}^{WA}}\right)\tensor[_{W}]{\ve{{{R}}}}{}^{{A}}\right)^{T}\exp\left({\bm{\delta}\bm{\theta}_{W}^{WB}}\right)\tensor[_{W}]{\ve{{{R}}}}{}^{{B}} \label{eq:rotation_pert}
\end{align}

\normalsize

Applying that\smalleq{ $\exp(\bm\delta\bm\theta) \approx \ve{I} + \mCross{\bm\delta\bm\theta}$} and \smalleq{$\mCross{\ve{a}}\ve{b} = -\mCross{\ve{b}}\ve{a}$} in \eqref{eq:translation_pert}; and \smalleq{$\ve{R}\exp(\bm\theta)\ve{R}^T = \exp(\ve{R}\bm\theta)$} and \smalleq{$\log(\exp(\bm\delta\bm\theta_1)\exp(\bm\delta\bm\theta_2))\approx \bm\delta\bm\theta_1 + \bm\delta\bm\theta_2$} in \eqref{eq:rotation_pert} we obtain the following expression for the covariance propagation:

\footnotesize

\begin{align}
\bm{\Sigma}_A^{AB} = 
  \tfrac{\bm{\partial}\tensor[_{A}]{\ve{{{T}}}}{}^{{B}}}{\bm{\partial}\tensor[_{W}]{\ve{{{T}}}}{}^{{A}}} \bm{\Sigma}_W^{WA}  \left(\tfrac{\bm{\partial}\tensor[_{A}]{\ve{{{T}}}}{}^{{B}}}{\bm{\partial}\tensor[_{W}]{\ve{{{T}}}}{}^{{A}}}\right)^T + 
  \tfrac{\bm{\partial}\tensor[_{A}]{\ve{{{T}}}}{}^{{B}}}{\bm{\partial}\tensor[_{W}]{\ve{{{T}}}}{}^{{B}}} \bm{\Sigma}_W^{WB} \left(\tfrac{\bm{\partial}\tensor[_{A}]{\ve{{{T}}}}{}^{{B}}}{\bm{\partial}\tensor[_{W}]{\ve{{{T}}}}{}^{{B}}}\right)^T
\end{align}

\normalsize

with

\scriptsize


\begin{align}
 \dfrac{\bm{\partial}\tensor[_{A}]{\ve{{{T}}}}{}^{{B}}}{\bm{\partial}\tensor[_{W}]{\ve{{{T}}}}{}^{{A}}} = \dfrac{\bm{\partial}(
   \bm{\delta}\ve{t}_A^{AB},\   \bm{\delta}\bm{\theta}_A^{AB})}{\bm{\partial}(
   \bm{\delta}\ve{t}_W^{WA},\   \bm{\delta}\bm{\theta}_W^{WA})}   
    = -\left(
 \begin{array}{cc}
 \tensor[_{A}]{\ve{{{R}}}}{}^{{W}} &  \tensor[_{A}]{\ve{{{R}}}}{}^{{W}}\mCross{\ve{t}_W^A-\ve{t}_W^B} \\[4pt]
 \ve{0} & \tensor[_{A}]{\ve{{{R}}}}{}^{{W}}
 \end{array}
 \right)
\end{align}


\begin{align}
 \dfrac{\bm{\partial}\tensor[_{A}]{\ve{{{T}}}}{}^{{B}}}{\bm{\partial}\tensor[_{W}]{\ve{{{T}}}}{}^{{B}}} = \dfrac{\bm{\partial}(
   \bm{\delta}\ve{t}_A^{AB},\   \bm{\delta}\bm{\theta}_A^{AB})}{\bm{\partial}(
   \bm{\delta}\ve{t}_W^{WB},\   \bm{\delta}\bm{\theta}_W^{WB})}   
    = \left(
 \begin{array}{cc}
 \tensor[_{A}]{\ve{{{R}}}}{}^{{W}} &  \ve{0} \\[4pt]
 \ve{0} & \tensor[_{A}]{\ve{{{R}}}}{}^{{W}}
 \end{array}
 \right)
\end{align}

\normalsize

\section{Pose-graph Jacobians and Information matrix} \label{app:posegraphjacs}

Frequently in optimisation problems, the measurement contribution to the error term is expressed through an addition or 
subtraction, and thus the information matrix of the constraint is trivially the inverse of the measurement's covariance. 
However in a pose-graph problem the error is expressed as: 

\small

\begin{align}
 \ve{e}_{ij} = \log_{\mathbb{SE}(3)}\left(\tensor[_{i}]{\ve{\hat{T}}}{}^{{j}}(\tensor[_{W}]{\ve{T}}{}^{{j}})^{-1}\tensor[_{W}]{\ve{T}}{}^{{i}}\right) \label{eq:error_posegraph_app}
\end{align}

\normalsize

and thus we need to compute the derivative of \smalleq{$\ve{e}_{ij}$} wrt. \smalleq{$\tensor[_{i}]{\ve{\hat{T}}}{}^{{j}}$}. Assuming that the covariance of \smalleq{$\tensor[_{i}]{\ve{\hat{T}}}{}^{{j}}$}, \smalleq{$\bm{\Sigma}_i^{ij}$}, is computed on the left side, we will compute the derivative applying a perturbation at the left of \smalleq{$\tensor[_{i}]{\ve{\hat{T}}}{}^{{j}}$}:

\footnotesize

\begin{align}
 \ve{e}_{ij} + \bm{\delta}\ve{e}_{ij} &= \log\left(
  \exp\right(\bm{\delta}\ve{t}_i^{ij}, \ \bm{\delta}\bm\theta_i^{ij}\left)\ve{E}_{ij}\right) \\[4pt]
  &\approx \left(
\begin{array}{c}
 \bm{\Delta}\ve{t}_{ij} - \mCross{\bm{\Delta}\ve{t}_{ij}}\bm{\delta}\bm\theta_i^{ij} + \bm{\delta}\ve{t}_i^{ij} \\[4pt]
 \log\left(\exp(\bm{\delta}\bm\theta_i^{ij})\bm{\Delta}\ve{R}_{ij}\right)
 \end{array}
\right)
\end{align}


\normalsize

\noindent with

\footnotesize 

\begin{align}
\ve{E}_{ij} = \tensor[_{i}]{\ve{\hat{T}}}{}^{{j}}(\tensor[_{W}]{\ve{T}}{}^{{j}})^{-1}\tensor[_{W}]{\ve{T}}{}^{{i}} = \left(
\begin{array}{cc}
 \bm\Delta\ve{R}_{ij} & \bm\Delta\ve{t}_{ij} \\
 \ve{0}^T & 1 
\end{array}
\right)
\end{align}

\normalsize

The derivative of the error wrt. the relative pose measurement remains:

\footnotesize

\begin{align}
 \frac{\partial\ve{e}_{ij}}{\partial\tensor[_{i}]{\ve{\hat{T}}}{}^{{j}}} = \frac{\partial\ve{e}_{ij}}{\bm{\partial}(
   \bm{\delta}\ve{t}_i^{ij},\   \bm{\delta}\bm{\theta}_i^{ij})} = \left(
  \begin{array}{cc}
  \ve{I} & -\mCross{\bm\Delta\ve{t}_{ij}} \\[4pt]
  \ve{0} & \bm{\Delta}\ve{Q}_{ij}^{-1}
  \end{array} \right)
\end{align}

\normalsize

\noindent where \smalleq{$\bm{\Delta}\ve{Q}_{ij} = \int_0^1\exp(\tau\log(\bm{\Delta}\ve{R}_{ij}))d\tau$}. And hence we obtain a expression for information update:

\footnotesize 

\begin{align}
 \bm{\Omega}_{ij} = \left(\frac{\partial\ve{e}_{ij}}{\partial\tensor[_{i}]{\ve{\hat{T}}}{}^{{j}}}\bm{\Sigma}_i^{ij}\left(\frac{\partial\ve{e}_{ij}}{\partial\tensor[_{i}]{\ve{\hat{T}}}{}^{{j}}}\right)^T\right)^{-1}
\end{align}

\normalsize

To obtain the Jacobians wrt. to the state of the optimised nodes we proceed similarly. The incremental updates are to be 
applied in the form of transformation matrices  by post-multiplication. Thus in this case the perturbations to compute 
the Jacobians are applied on the right side of the absolute camera poses in \eqref{eq:error_posegraph_app}. Thus we 
obtain:

\footnotesize 

\begin{align}
 \frac{\partial\ve{e}_{ij}}{\partial\tensor[_{W}]{\ve{\hat{T}}}{}^{{i}}} = \frac{\partial\ve{e}_{ij}}{\bm{\partial}(
   \bm{\delta}\ve{t}_i^{Wi},\   \bm{\delta}\bm{\theta}_i^{Wi})} = \left(
  \begin{array}{cc}
  \bm{\Delta}\ve{R}_{ij} & \ve{0} \\[4pt]
  \ve{0} & \bm{\Delta}\ve{Q}_{ij}^{-1}\bm{\Delta}\ve{R}_{ij}
  \end{array} \right)
\end{align}

\normalsize

\footnotesize 

\begin{align}
 \frac{\partial\ve{e}_{ij}}{\partial\tensor[_{W}]{\ve{\hat{T}}}{}^{{j}}} = \frac{\partial\ve{e}_{ij}}{\bm{\partial}(
   \bm{\delta}\ve{t}_j^{Wj},\   \bm{\delta}\bm{\theta}_j^{Wj})} = \left(
  \begin{array}{cc}
  -\tensor[_{i}]{\ve{\hat{R}}}{}^{{j}} & \mCross{\ve{\hat{t}}_{ij}}\tensor[_{i}]{\ve{\hat{R}}}{}^{{j}} \\[4pt]
  \ve{0} & -\bm{\Delta}\ve{Q}_{ij}^{-1}\tensor[_{i}]{\ve{\hat{R}}}{}^{{j}}
  \end{array} \right)
\end{align}

\normalsize

\bibliographystyle{plain}

%
%

%




\end{document}